\documentclass{article}
\usepackage[dblblindworkshop,final]{neurips_2026}
\workshoptitle{Diffusion Language Models: Foundations, Efficiency, and Reasoning (DiffuLM)}
\usepackage[utf8]{inputenc}
\usepackage[T1]{fontenc}
\usepackage{amsmath,amssymb,amsthm,mathtools}
\usepackage{xfrac}
\usepackage{microtype}
\usepackage{booktabs}
\usepackage{array}
\DeclareMathOperator{\arctanh}{arctanh}
\usepackage[shortlabels]{enumitem}
\usepackage{hyperref}
\hypersetup{pdftitle={End-to-End Guarantees for Masked Diffusion Models under Conditional Mixing},pdfauthor={Fareed Sheriff}}
\usepackage[capitalize,noabbrev]{cleveref}
\usepackage{pgfplots}\pgfplotsset{compat=1.18}\usepgfplotslibrary{fillbetween}

\theoremstyle{plain}
\newtheorem{theorem}{Theorem}
\newtheorem{lemma}{Lemma}
\newtheorem{proposition}{Proposition}
\newtheorem{corollary}{Corollary}
\theoremstyle{definition}
\newtheorem{assumption}{Assumption}
\newtheorem{definition}{Definition}
\theoremstyle{remark}
\newtheorem{remark}{Remark}
\newtheorem*{remark*}{Remark}
\crefname{assumption}{Assumption}{Assumptions}
\Crefname{assumption}{Assumption}{Assumptions}

\newcommand{\cV}{\mathcal{V}}
\newcommand{\cB}{\mathcal{B}}
\newcommand{\cVm}{\overline{\mathcal{V}}}
\newcommand{\lmax}{\ell_{\max}}
\newcommand{\kpi}{\kappa}
\newcommand{\tauc}{\tau_c}
\newcommand{\Czero}{C_0}
\newcommand{\Efact}{\mathcal E_{\mathrm{fact}}}
\newcommand{\Elearn}{\mathcal E_{\mathrm{learn}}}
\newcommand{\Prob}{\mathbb{P}}
\newcommand{\E}{\mathbb{E}}
\DeclareMathOperator{\KL}{KL}
\DeclareMathOperator{\supp}{supp}
\DeclareMathOperator{\TV}{TV}
\DeclareMathOperator{\TC}{TC}
\newcommand{\Unif}{\mathrm{Unif}}
\newcommand{\etamod}{\eta}
\renewcommand{\O}{\mathcal{O}}

\newcommand{\eps}{\varepsilon}
\newcommand{\bbZ}{\mathbb{Z}}
\newcommand{\dH}{d_{\mathrm H}}
\newcommand{\pdiam}{\Delta_{\mathrm H}}
\newcommand{\cH}{\mathcal{H}}
\newcommand{\indep}{\perp \!\!\! \perp}
\newcommand{\eD}{\gamma_{\mathrm{D}}}
\newcommand{\mask}{\mathsf{[mask]}}

\begin{document}
\title{End-to-End Guarantees for Masked Diffusion Models under Conditional Mixing}
\author{Fareed Sheriff\\Massachusetts Institute of Technology\\\texttt{fareeds@mit.edu}}
\maketitle

\begin{abstract}
	Masked diffusion models train a denoiser on masked cross-entropy and generate by unmasking many positions in parallel. Existing MDM sampling bounds assume an exact denoiser or leave its error uncontrolled while training bounds cover only the masking distribution on which the model was trained. We present an end-to-end guarantee for a \textit{learned} denoiser under conditional mixing (CM): stationary sources (e.g. floored order-$r$ Markov and two-sided-Doeblin hidden-Markov sources) for which revealing distant visible tokens moves a masked token's posterior on average exponentially little in the distance with correlation length $\tauc$. We show the empirical risk minimizer trained under any masking distribution attains per-token excess $\eps'$ on that distribution from $\eps'^{-\O(\tauc\log|\cV|)}$ training sequences, independent of sequence length $N$ --- one sample budget serves every decoding schedule when each schedule's denoiser is trained under its own weighting. We further show that the exponent's order $\Theta(\tauc\log|\cV|)$ is minimax-optimal up to constants; more efficient rates require added structure. With this, we bound the end-to-end KL as $\KL(\pi\Vert p_{\mathrm{alg}})\leq \eps+N\eps'$ under a dilated schedule of $T=\O(\tauc\log(N|\cV|/\eps))$ denoising steps versus $\Omega(N)$ steps for a fixed random order on an explicit CM source. Finally, we show that block-autoregressive samplers chain across blocks with no added error and that a $\Lambda$-looped order-$r$ message-passing class on order-$r$ sources achieves polynomial sample complexity, with approximation error improving per loop until saturation at $\Lambda r\approx \tau_c\log(1/\eps')$. Experiments on released and retrained models support our results' qualitative predictions.
\end{abstract}

\section{Introduction}\label{sec:intro}

\par Masked diffusion language models (MDMs) trade the autoregressive factorization for per-step conditional independence between tokens, which lets them generate many tokens in parallel. MDMs use an absorbing-state/masked forward process on tokens \citep{austin2021,campbell2022,shi2024,sahoo2024} and generate by unmasking several positions per step. Discrete diffusion training theory and sampling theory have largely developed separately. On the sampling side, recent work bounds the convergence and step count of the reverse process \citep{zhangchengu2024,chenying2024,liang2025discrete,liang2026sharp}, gives information-theoretic accounts of the factorization error \citep{licai2025,lavenant2025,dmitriev2026}, limits for parallel generation \citep{chen2025optimal}, and analyses of confidence-adaptive decoding \citep{cai2026confidence}. Every result assumes an exact denoiser or leaves the denoiser's error at the masking patterns that decoding visits as a free parameter. On the training side, finite-sample bounds exist for generative masked language modeling \citep{li2024mple} and discrete diffusion score estimation \citep{srikanth2026,cho2026minimax}; these control the learned model at the masking distribution on which it was trained and compose it with a matched sampler (Gibbs-style chains or CTMC samplers). Without structural assumptions, their rates scale with the full state space $|\cV|^N$ \citep{cho2026minimax}. Composing training with the parallel-sampling decompositions above requires a bound covering the masking patterns of an \emph{arbitrary} decoding strategy with a per-token sample budget independent of the strategy and $N$; no existing result provides one.

\par We provide such a bound for stationary data whose token-to-token influence on average decays exponentially with distance, a source family we call conditional mixing (\cref{ass:cm}), which covers finite-order Markov sources with a transition floor (e.g. smoothed $n$-grams) and hidden Markov sources whose hidden kernel satisfies a two-sided Doeblin condition. We show that under conditional mixing (CM),
\begin{itemize}
	\item \emph{Training}: the empirical risk minimizer trained under any masking distribution $\omega$ has per-token excess $\eps'$ at $\omega$ with $(1/\eps')^{\O(\tauc\log{|\cV|})}$ training sequences, independent of sequence length. We further show this is minimax-optimal in the exponent's order $\tauc\log|\cV|$ (correlation length times log-vocabulary size).
	\item \emph{End-to-end \& dilated decoding}: combining this with the training-sampling decomposition of \citet{lavenant2025}, we show $\KL(\pi\Vert p_{\mathrm{alg}})\le\Efact(\nu)+N\eps'$ for any non-adaptive decoding strategy $\nu$ whose denoiser is trained under $\nu$'s induced weighting. A dilated schedule with spacing $b= \O(\tau_c\log(N|\cV|/\eps))$ has $\Efact\le\eps$ in $T=b$ steps, giving $\eps+N\eps'$; a fixed random order needs $\Omega(N)$ steps on a two-state Markov chain.
	\item \emph{Architectures}: block-autoregressive samplers chain across blocks with no added error, and on floored order-$r$ sources a looped order-$r$ message-passing class with loop count in $\O((\tauc/r)\log(1/\eps'))$ achieves polynomial sample complexity $\tilde O(|\cV|^{r+1}/\eps'^2)$.
	\item \emph{Experiments}: we probe the influence assumptions on a released MDM, block size on a synthetic source, and dilated versus random decoding on released and finetuned models (\cref{sec:empirics}).
\end{itemize}
Proofs and experimental details are in the appendices.

\section{Related work}\label{sec:related}

\paragraph{Learning the denoiser.} The learning results cited in \cref{sec:intro} hold at the training masking distribution. Our bound holds at every masking distribution with one per-token budget that does not depend on $N$, provided the denoiser is trained under the decoding schedule's weighting (\cref{thm:upper}). Its window argument extends short-memory prediction \citep{sharan2018prediction} from one-sided to two-sided conditionals under arbitrary masks. Its lower bound (\cref{thm:minimax}) parallels minimax rates for predicting Markov chains \citep{hao2018learning,han2021optimal} and hidden Markov models \citep{han2024prediction}.

\paragraph{Composing learning with sampling.} We use the decomposition of \citet{lavenant2025}, which splits the sampler's KL into a learning term and the factorization error. Earlier end-to-end results pair the learned model with a sampler matched to its training distribution \citep{li2024mple,cho2026minimax}. \Cref{cor:endtoend} pairs it with any non-adaptive decoding strategy, including parallel ones, and with adaptive strategies under strong conditional mixing.

\paragraph{Parallel decoding.} With exact conditionals, the KL of a deterministic adaptive decoding policy equals its accumulated conditional total correlation \citep{benhamu2025}, and order-$r$ Markov chains admit exact schedules with $r(\lceil\log_2(N/r)\rceil+1)$ steps \citep{wen2026conditional}. For random walks on graphs, a bisection sampler is exact in logarithmically many steps \citep{bansal2026parallel}. On order-$r$ sources our dilated schedule uses $\O(\tauc\log N)$ steps with $\tauc\le r/(2q_{\min}^r)$ (\cref{prop:markov}). It needs only conditional mixing, which includes Doeblin hidden Markov sources, and it works with a learned denoiser. Under Dobrushin-type conditions, parallel Markov chain Monte Carlo samples Gibbs distributions in $\O(\log N)$ parallel rounds \citep{liu2022simple}, and the LOCAL algorithm of \citet{feng2017what} needs an extra factor of the maximum degree.

\paragraph{Mixing conditions.} Conditional mixing adapts strong spatial mixing \citep{weitz2006} to masked prediction and averages it over the source (\cref{sec:cm}). Our hidden Markov result (\cref{prop:hmm}) extends filter stability under a mixing kernel \citep{legland2004,atarzeitouni1997} to two-sided conditionals under arbitrary masks.

\section{Masked diffusion and conditional mixing}\label{sec:cm}

\par Let $\cV$ be a finite vocabulary, $\cVm=\cV\cup\{\mask\}$ the vocabulary augmented with the mask token $\mask$, $z\subseteq[N]$ the set of visible (unmasked or previously predicted) positions, $M=[N]\setminus z$ the masked positions, and $\pi$ a distribution on $\cV^N$ from which training sequences are drawn i.i.d. We assume throughout this work that $\pi$ is the length-$N$ marginal of a stationary stochastic process.%

\paragraph{Forward process and objective} Masked or absorbing-state diffusion corrupts tokens independently and irreversibly: at time $t\in [0,1]$ each token survives with probability $\alpha_t$ and is replaced by $\mask$ otherwise \citep{austin2021,sahoo2024,shi2024}. We analyze the unweighted masked cross-entropy
\begin{equation}\label{eq:ce}
	\mathcal{L}_{\mathrm{CE}}(f) = -\E_{t\sim U[0,1]}\E_{x,z}\left[\frac{1}{|M|}\sum_{i\in M}{\log{f(x_i\mid i,z,x_z)}}\right],
\end{equation}
where the \emph{denoiser} $f$ maps each masked position and partially-masked sequence to a distribution over $\cV$, and the empty-mask event contributes zero. The denoiser needs no time input: its targets are the time-independent conditionals $\pi(x_i\in\cdot\mid x_z)$ \citep{ou2025,zheng2024}.

\paragraph{Risk and weightings} Denote the risk $R_\omega(f)=\E_\omega[-\log f(x_i\mid i,z,x_z)]$ where $(i,z,x)$ is drawn from a joint law $\omega$ (a \emph{weighting}) with $x\sim\pi$, and excess risk $\mathcal E_\omega(f)=R_\omega(f)-R_\omega(\pi)$ over the true conditionals. Call $\omega$ \emph{admissible} if $x_i\mid(i,z,x_z)\sim\pi(\cdot\mid x_z)$, so that
\begin{equation}\label{eq:excessKL}
	\mathcal E_\omega(f)=\E_\omega\left[\KL\left(\pi(\cdot\mid x_z)\Vert f(\cdot\mid i,z,x_z)\right)\right]\geq0;
\end{equation}
every masking distribution (where $(i,z)$ is independent of $x$) is admissible, and so is the weighting $\omega_\nu$ induced by running any decoding strategy $\nu$ on $x\sim\pi$ under teacher forcing (\cref{prop:adaptive}).

\paragraph{Training as risk minimization} Let the uniform masking distribution $\omega_{\mathrm u}$ draw $i$ uniformly and $z$ as a uniform subset of $[N]\setminus\{i\}$ of uniform size. Then with noising schedule $\alpha_t=1-t$,
\begin{equation}\label{eq:pretrainid}
	\mathcal{L}_{\mathrm{CE}}(f) = \frac{N}{N+1}R_{\omega_{\mathrm u}}(f)
\end{equation}
and training on masked CE \citep{sahoo2024} is empirical risk minimization under $\omega_{\mathrm u}$, the objective-level equivalence with any-order autoregression (\citealp{uria2014,hoogeboom2022,ou2025,shi2024}); other schedules reweight $\omega_{\mathrm u}$.%

\paragraph{Sampler} A monotone sampler (Algorithm 1 of \citealp{lavenant2025}) unmasks sets of positions over $T$ steps, filling each commit set $S$ conditionally independently given the visible tokens. The stepwise error splits exactly (cf. \citealp{benhamu2025}) as
\begin{equation}\label{eq:stepsplit}
\KL\left(\pi(x_S\mid x_z)\Big\Vert\prod_{i\in S}
f(\cdot\mid i,z,x_z)\right)
=\sum_{i\in S}\KL\left(\pi(x_i\mid x_z)\Vert
f(\cdot\mid i,z,x_z)\right)
+\TC_\pi\left(x_S\mid x_z\right),
\end{equation}
where the second term is the total correlation of the true conditional, untouched by training. Existing theory bounds the total correlation summed over steps, the \emph{factorization error}, by global functionals (mutual-information couplings \citep{licai2025}, the information profile \citep{lavenant2025}, effective total correlation \citep{dmitriev2026}) that neither localize it nor address learning. Throughout, we bound the per-token excess risk by $\eps'$ and the factorization error by $\eps$, so the sampler's output law $p_{\mathrm{alg}}$ satisfies $\KL(\pi\Vert p_{\mathrm{alg}})\le\eps+N\eps'$ \citep{lavenant2025}.%

\paragraph{Conditional mixing} To compose training and sampling guarantees, we need a source assumption that both limits effective model dimension (so the denoiser is learnable from finitely many samples) and bounds the total correlation among co-committed tokens (so the factorization error per sampling step is controlled). We adapt strong spatial mixing \citep{weitz2006}, which Dobrushin's uniqueness condition \citep{dobrushin1968} implies, to masked prediction. The averaged influence coefficient $\bar\kappa(d)$ measures how far revealing visible tokens beyond distance $d$ moves a masked token's posterior on average over the source, and conditional mixing requires it to decay exponentially in $d$:

\begin{assumption}[Conditional mixing]\label{ass:cm}
	Define the \emph{averaged influence coefficient}
\begin{equation}\label{eq:kappaavg}
	\bar\kappa(d) = \sup_{\substack{i, z\not\ni i,\\ A\subseteq z\setminus\cB(i,d)}}\E_{x\sim\pi}\TV\left(\pi(x_i\in\cdot\mid x_z), \pi(x_i\in\cdot\mid x_{z\setminus A})\right)
\end{equation}
where $\cB(i,d)=\{j:|j-i|\le d\}$. Distribution $\pi$ satisfies \emph{conditional mixing} (CM) with correlation length $\tauc>0$ and prefactor $C_0\geq 1$ if
\begin{equation}
	\bar\kappa(d) \leq C_0e^{-d/\tauc}\text{ for all }d\geq 0
\end{equation}
\end{assumption}

\par The worst-case form of $\bar\kappa$, the \emph{influence coefficient}, compares pairs of visible configurations that agree near $i$:
\begin{equation}\label{eq:kappa}
	\kappa(d) = \sup_{\substack{i, z\not\ni i,\\ \xi\sim_d\xi'}}\TV\left(\pi(x_i\in\cdot\mid x_z=\xi), \pi(x_i\in\cdot\mid x_z=\xi')\right)
\end{equation}
where $\xi\sim_d\xi'$ means $(\xi,\xi')\in\cV^z\times\cV^z$ agree on $z\cap\cB(i,d)$ and both have positive probability under $\pi$.
\begin{assumption}[Strong conditional mixing]\label{ass:strong}
$\pi$ satisfies \emph{strong conditional mixing} with correlation length $\tauc>0$ and prefactor $C_0\geq 1$ if $\kappa(d)\leq C_0e^{-d/\tauc}$ for all $d\geq 0$.
\end{assumption}

\par Conditional mixing is \emph{weaker} than conditional independence: it permits dependence among tokens but bounds the sampler's factorization error; tokens committed $b$ apart carry zero factorization error whenever $\bar\kappa(b-1)=0$, by \eqref{eq:factbound}. Strong CM implies CM with the same constants, since $\bar\kappa\le\kpi$ by convexity of total variation \eqref{eq:mixtv}, but not conversely: CM tolerates configurations with large influence if they are rare under $\pi$. Strong CM (and thus CM) contains the classical autoregressive families with constants independent of sequence length (proofs in \cref{app:certificates}):

\begin{proposition}[Order-$r$ Markov sources]\label{prop:markov}
Draw training sequences from a stationary order-$r$ Markov source on $\cV$ whose transition matrix $Q$ has floor $q_{\min}=\min_{a\in\cV,y\in\cV^r}Q(a\mid y)>0$ and let $\Delta_r=2r\log(1/q_{\min})$ and $\vartheta_r=\tanh(\Delta_r/4)=(1-q_{\min}^r)/(1+q_{\min}^r)$. Then, the data distribution satisfies \cref{ass:strong} with $\tauc=r/\log(1/\vartheta_r)\le r/(2q_{\min}^r)$ and $\Czero=2\Delta_r/\vartheta_r^3$.
\end{proposition}

\begin{proposition}[Hidden Markov sources with a Doeblin kernel]\label{prop:hmm}
Let $(h_t)_{t\in\bbZ}$ be a stationary Markov chain on finite set $\cH$ whose kernel $\Gamma$ satisfies for some $\eD\in(0,1)$ and some measure $\nu_{\mathrm D}$ on $\cH$ the two-sided Doeblin condition
\begin{equation}\label{eq:AZ}
	\eD\nu_{\mathrm D}(h')\le \Gamma(h,h')\leq \eD^{-1}\nu_{\mathrm D}(h')\forall h,h'\in\cH
\end{equation}
and let the observations $x_t\sim g(\cdot\mid h_t)$ be conditionally independent given $(h_t)_{t\in\bbZ}$ for any emission kernel $g$, $\Delta_\Gamma=4\log(1/\eD)$, and $\vartheta_\Gamma=\tanh(\Delta_\Gamma/4)=(1-\eD^2)/(1+\eD^2)$.

Then, $(x_t)_{t\in\bbZ}$ satisfies \cref{ass:strong} with $\tauc=1/\log(1/\vartheta_\Gamma)\le1/(2\eD^2)$ and $\Czero=2\Delta_\Gamma/\vartheta_\Gamma$.
\end{proposition}

\par Long-range parity constraints force $\kpi\equiv1$ and $\bar\kappa\ge1/2$, as in known hard instances for parallel sampling and masked learning \citep{chen2025optimal,kim2025}. Conditional mixing is nonetheless a strong requirement: taking a single visible token in \cref{ass:cm} bounds one token's average influence on another with \emph{nothing} between them visible so pairwise mutual information decays exponentially; in text, pairwise MI decays as a power law \citep{lin2017,shen2019mi,nakaishi2025}, which CM excludes (\cref{rem:powerlaw}); with context visible, however, our probe of a trained MDM (\cref{sec:empirics}) finds influence decaying exponentially on average but not in the worst case, which motivates CM's averaged form over strong CM. We assume CM for most of our proofs and state where we assume strong CM.

\section{Learning the denoiser}\label{sec:learning}

\par Training minimizes the masked cross-entropy \eqref{eq:ce} (risk at $\omega_{\mathrm u}$), but decoding queries the denoiser at masking patterns a training-time analysis need not cover. The bound below is \emph{schedule-uniform in the sample budget}: the same $n$ suffices at every masking distribution $\omega$, with the ERM taken under that $\omega$. It does not cover a model trained at one weighting and evaluated at another (the mask-training design question of \citealp{shih2022}).%

\par The denoising hypothesis class $\mathcal F_w$ consists of denoisers that read only the $w$ positions to each side of the masked target, outputting a distribution over $\cV$ with every probability at least $p_{\min}$, keeping the log-loss below $\lmax=\log(1/p_{\min})$ (\cref{def:window}). Conditional mixing makes a window of width $w\asymp\tauc\log(1/\varepsilon')$ sufficient ($\log|\cVm|$ and $\log|\cV|$ agree to leading order).

\begin{theorem}[Sample complexity of masked-diffusion training]\label{thm:upper}
	Let $\pi$ be stationary (\cref{sec:cm}), $N\ge2$, $\varepsilon'\in(0,1]$ and $\delta\in(0,1)$.
	\begin{enumerate}[(a),nosep]
	\item If \cref{ass:cm} holds with correlation length $\tauc$ and prefactor $\Czero$, then for $p_{\min}=\eps'/(10|\cV|\log(10|\cV|/\eps'))$, $w=\lceil\tauc\log(4\Czero(\lmax+1)/\varepsilon')\rceil$ and every masking distribution $\omega$, the ERM $\hat f$ over $\mathcal F_w$ trained under $\omega$ satisfies $\mathcal E_\omega(\hat f)\le\eps'$ w.p.\ $\geq 1-\delta$ whenever
	\begin{equation*}
		n \geq \left(1/\eps'\right)^{2+2\tauc\log{|\cVm|}+o(1)}|\cV|\log\frac{2}{\delta},
	\end{equation*}
	the $o(1)$ taken as $\eps'\to0$ at fixed $(\tauc,\Czero,|\cV|)$.
	\item If \cref{ass:strong} holds with $(\tauc,\Czero)$, (a) holds for every admissible $\omega$, including the weightings induced by adaptive decoding strategies fixed before the training sample is drawn (\cref{prop:adaptive}).
	\end{enumerate}
\end{theorem}
\emph{Proof sketch.} We can decompose the excess of $\hat{f}$ into an approximation error and estimation error. Specifically, the excess is at most the excess of the best denoiser in $\mathcal F_w$ (the approximation error) plus the cost of choosing a denoiser from $n$ samples (the estimation error). For the first, we use the true conditional given only visible tokens within distance $w$ mixed with a uniform so every probability is at least $p_{\min}$; then, conditional mixing (strong conditional mixing in (b)) bounds its expected TV from the true conditional by $\Czero e^{-w/\tauc}$ plus the mixing weight, and the floor $p_{\min}$ turns a total variation $t$ into a KL of at most $(\lmax+1)t$.
For the second, any element of $\mathcal F_w$ is a distribution over $\cV$ for each of the $|\cVm|^{2w}$ possible window contents (a token or the mask at each of the $2w$ positions); Hoeffding with union bound over a grid of these $|\cVm|^{2w}|\cV|$ probabilities gives an error of order $\lmax\sqrt{|\cVm|^{2w}|\cV|\log(1/\eps')/n}$. The stated $w$ makes the first error at most $\eps'/2$ and the stated $n$ the second (\cref{app:learning}).
\par Window truncation is a common idea in one-sided prediction \citep{csiszartalata2006}; we extend it in \cref{thm:upper} to two-sided conditionals under arbitrary masks.%

\paragraph{Minimax optimality and complexity hierarchy} The order of the exponent in \cref{thm:upper} cannot be improved:
\begin{theorem}[Minimax lower bound]\label{thm:minimax}
	Let $\tauc\ge1$, $\Czero\ge16$ and $|\cV|\ge2$. Then, there is an explicit $\eps_0=\eps_0(\tauc,\Czero)>0$ such that, for $\eps'\in(0,\eps_0]$ and $d$ the largest integer that satisfies $160d\exp({3d/\tauc})\sqrt{\eps'}\le\Czero$, there is for every $N\geq 2d$ a finite family $\Pi$ of stationary sources on $\cV^{\bbZ}$ with uniform marginals each satisfying \cref{ass:strong} with $(\tauc,\Czero)$ such that for the weighting $\omega^*$ that masks the final token and reveals only the $d$ tokens preceding it every map $\hat f$ from $n$ unmasked training sequences to a predictor satisfies
\begin{equation*}
	\max_{\pi\in\Pi}\E_\pi\left[\mathcal E_{\omega^*}(\hat f)\right]\ge\eps'\qquad\text{for}\qquad n(N-d)<\frac{|\cV|^{d-1}}{300\eps'}=\left(\frac1{\eps'}\right)^{1+(1-o(1))\frac{\tauc}6\log|\cV|}.
\end{equation*}
\end{theorem}
\emph{Proof sketch.} The sources are order-$d$ Markov chains whose next-token distribution is uniform up to a perturbation of relative size $\gamma\asymp\sqrt{\eps'}$ with sign chosen separately for each of the $|\cV|^{d-1}$ values of the preceding $d-1$ tokens. This perturbation is sufficiently small that every such chain satisfies strong conditional mixing with $(\tauc,\Czero)$. Then, predicting the final token at $\omega^*$ with excess below $\eps'$ requires most of the signs and a sign can be learned only from tokens that follow its $d-1$ preceding values all of which occurs w.p. $|\cV|^{-(d-1)}$ per token, and each sign needs about $1/\gamma^2\asymp1/\eps'$ such tokens; the sample therefore needs about $|\cV|^{d-1}/\eps'$ tokens. For $N\in\Theta(d)$, this is  $(1/\eps')^{1+\Omega(\tauc\log|\cV|)}$ training sequences, matching \cref{thm:upper} in exponent order. %
\par For text, the exponent is in the hundreds even at the local decay scale ${\approx}26$ with the GPT-2 vocabulary ($\log|\cVm|\approx10.8$); rates whose exponent does not grow with $\tauc\log|\cV|$ need added structure. Under a conditional floor $q_c$ on the masked conditionals (order-$r$ sources give $q_c\ge q_{\min}^{r+1}$) and CM in mean square, the TV-to-KL conversion becomes quadratic and the exponent's $\tauc$ term halves, yielding $2+\tauc\log|\cVm|+o(1)$ (\cref{cor:floored}). For order-$r$ Markov sources, the true conditional can be computed by iterating a transition kernel which gives $\tilde{O}(|\cV|^{r+1}/{\eps'}^2)$ (\cref{cor:looped}).%

\paragraph{Training objectives} Unweighted masked CE and the ELBO both estimate the risk at $\omega_{\mathrm u}$ (\cref{app:bridge}); other noise schedules and time weightings change $\omega$ \citep{sahooscaling2026} and thus which decoding strategies are covered. \Cref{thm:upper} covers the per-sequence ERM on a single masked token and the unweighted masked CE minimizer, which averages over all masked positions of a sequence; both per-sequence losses are bounded by $\lmax$.%

\section{End-to-end guarantees}\label{sec:orderings}

\par We compose \cref{thm:upper} with the training-sampling decomposition of \citet{lavenant2025}:
\begin{equation}\label{eq:lz-decomp}
	\KL(\pi\Vert p_{\mathrm{alg}})\leq \Elearn + \Efact,\qquad \Efact = \E\sum_{k\geq 1}\TC_{\pi}(x_{S_k}\mid x_{z_k})
\end{equation}
where $S_k$ is the $k$-th commit set and $z_k=S_1\cup\dots\cup S_{k-1}$ the visible set before it, $\Elearn$ is the denoiser's per-token excess at the weighting the strategy induces, accumulated over the $N$ commits, and $\Efact$ is the information cost of committing tokens in parallel. For a strategy $\nu$, \cref{thm:upper} applied to the ERM trained under $\omega_\nu$ (\cref{prop:adaptive}) controls $\Elearn$; \cref{cor:endtoend} states the composition and computes the constants for the dilated (residue-class) schedule, a constant-spacing relative of the coarse-to-fine dilated scheduler of \citet{luxembourg2025}. Empirical schedule design pursues the same step-count reduction \citep{park2024jump,punt2026}, \citet{peng2025path} plan which tokens to update, and \citet{zhao2026adaptation} adapt step sizes to intrinsic dependence.

\begin{corollary}[End-to-end guarantee]\label{cor:endtoend}
	Let $\pi$ be stationary and satisfy \cref{ass:cm}.
	\begin{enumerate}[(i),nosep]
		\item \emph{Composition.} Fix a non-adaptive decoding strategy $\nu$ (any $\nu$ under \cref{ass:strong}), chosen independently of the training sample, with induced weighting $\omega_{\nu}$ (\cref{prop:adaptive}). For $\hat{f}$ the ERM of \cref{thm:upper} trained under $\omega_{\nu}$, running Algorithm~1 with $(\nu,\hat{f})$ satisfies, w.p.\ at least $1-\delta$, $\KL(\pi\Vert p_{\mathrm{alg}})\leq \Efact(\nu) + N\eps'$.
\item \emph{Dilated factorization.} The dilated schedule that unmasks $\{t,t+b,t+2b,\dots\}$ at step $t$ achieves $\Efact\leq \eps$ at spacing $b=\O(\tau_c\log(C_0N|\mathcal{V}|/\eps))$ in $T=b$ steps, so part~(i) gives $\KL(\pi\Vert p_{\mathrm{alg}})\le\varepsilon+N\varepsilon'$ w.p. $\geq 1-\delta$.
	\end{enumerate}
\end{corollary}
\emph{Proof sketch for (ii).} Under a dilated schedule, tokens committed in the same step are at least $b$ apart. By chain rule, the factorization error is a sum over the $N-b$ tokens not first in their step of the MI between the token and tokens committed before it in the same step, given those already visible. Each term is the average drop in the token's conditional entropy when tokens at distance at least $b$ are revealed --- conditional mixing bounds this change of the conditional by $\bar\kappa(b-1)\le\Czero e^{-(b-1)/\tauc}$ in expected total variation; applying an inequality between the difference in entropy and the total variation described in the full proof, we show the term is at most $\etamod(\Czero e^{-(b-1)/\tauc})$ for some concave $\etamod$ we construct. The sum $(N-b)\etamod(\Czero e^{-(b-1)/\tauc})$ is then at most $\eps$ for $b\approx\tauc\log(\Czero N|\cV|/\eps)$ (\cref{app:orderings}).
\par \Cref{thm:upper}(a) and \cref{cor:endtoend,cor:blockar} use \cref{ass:cm} only through two schedule-averaged quantities: the truncation error $\bar\kappa_\omega(w)$ at the training weighting (\cref{lem:approx}) and the influence $\bar\kappa_\nu(b-1)$ between tokens committed in the same dilated step \eqref{eq:kappabar}. These results hold whenever the two averages obey the bound of \cref{ass:cm}, which is the averaged behavior our measurements probe (\cref{sec:empirics}).

\begin{remark}[Limits of averaging]\label{rem:powerlaw}
At spacing $b$ the first dilated step commits $\{1,1+b,\dots\}$ with nothing visible, so by the chain rule and stationarity $\Efact\ge(\lceil N/b\rceil-1)\,I(x_0;x_b)$. If $I(x_0;x_b)\ge cb^{-\alpha}$, as reported for text, then $\Efact\le\eps$ forces $b^{1+\alpha}\ge cN/(2\eps)$ for $N\ge2b$: polynomially many steps, and \cref{ass:cm} fails however small the influence is \emph{with} context.
\end{remark}

\paragraph{Random-order separation} \Cref{cor:endtoend}(ii) needs $\O(\tauc\log N)$ dilated steps; a fixed random order cannot match this on an explicit CM source:

\begin{corollary}[Random-order separation]\label{cor:randorder}
Let $\pi_q$ be the stationary two-state Markov chain with flip probability $q\in(0,\tfrac12)$ ($\tauc\le1/(2q)$ by \cref{prop:markov}). A \emph{random-order schedule} draws its step sizes independently of $x$ and commits positions in a uniformly random order $\sigma$; both are drawn once and then fixed, so each run is a deterministic schedule with error $\KL(\pi\Vert p^{\sigma})$. With $c_q=8q^4(1-q)^4(1-2q)^2$, for all $N\ge2$ and $\eps>0$, if $\E\,\KL(\pi\Vert p^{\sigma})\le\eps$ (over $\sigma$ and the sizes) with the exact conditionals, the expected number of steps satisfies
\begin{equation*}
	\E[T]\ge\frac{c_q\,N}{c_q+2\eps}.
\end{equation*}
In particular $\E[T]=\Omega(N)$ at fixed $\eps$, where dilated needs $\O(\tauc\log(N/\varepsilon))$.
\end{corollary}
\emph{Proof sketch.} With exact conditionals, the KL equals the factorization error. A step that commits $g$ tokens contains $g(g-1)/2$ pairs of tokens committed together; under uniform order, any pair are neighbors with probability $\Theta(1/N)$ so each pair adds at least $c_q/N$ (where $c_q$ is the increase in KL due to a pair of neighbors). With step sizes $g_1,\dots,g_T$, the expected KL is at least $\frac{c_q}{2N}\sum_kg_k(g_k-1)\ge\frac{c_q}2\left(\frac NT-1\right)$, where $\sum_kg_k=N$ forces $\sum_kg_k^2\ge N^2/T$ (\cref{app:orderings}).
\par \citet{feng2025theoretical} earlier proved an $\Omega(N)$ random-order lower bound for a sequence-error metric on a local copying source; \cref{cor:randorder} is a per-order KL version with explicit constants.

\section{Architecture bounds}\label{sec:arch}

\subsection{Block-autoregressive samplers}\label{sec:blockar}

\par Block-autoregressive generation partitions a sequence into blocks of size $B$, generating blocks left to right autoregressively and running masked diffusion within each block given the generated prefix \citep{arriola2025}; recent works using block diffusion include LLaDA2.0 \citep{llada2025scale} and DiffusionGemma \citep{diffusiongemma2026}, which uses uniform noise instead of masking. KL decomposes exactly over blocks, and the within-block weightings mix into one masking distribution, so \cref{thm:upper} applies once.

\begin{corollary}[Block-autoregressive samplers]\label{cor:blockar}
Let $\pi$ be stationary and satisfy \cref{ass:cm}, $N=mB$, partition $[N]$ into consecutive blocks of size $B$,
and generate blocks left to right, each by Algorithm~1 on its block
with the generated prefix revealed and an $x$-independent
within-block strategy.
	\begin{enumerate}[(i),nosep]
	\item Train the denoiser under this sampler's induced weighting (constructed in \cref{app:orderings}) at the same $n$ as \cref{thm:upper}. Then w.p. $\geq 1-\delta$,
\begin{equation*}
	\KL(\pi\Vert p_{\mathrm{alg}})\leq {\underbrace{\sum_{k=1}^{m}\E_\pi\left[\Efact^{(k)}\right]}_{\Efact^{\mathrm{blk}}}} + N\eps',
\end{equation*}
where $\Efact^{(k)}$ is the factorization error of block $k$'s run given its prefix.
\item The dilated schedule at spacing $\min\{b^*(\eps),B\}$ within each block achieves $\Efact^{\mathrm{blk}}\le\eps$ in
\begin{equation*}
	T=\frac NB\cdot\min\{b^*(\eps),B\}\ \text{ steps,}
\end{equation*}
where $b^*(\eps)=\O(\tauc\log(\Czero N|\cV|/\eps))$ comes from \cref{cor:endtoend}(ii).
\end{enumerate}
\end{corollary}

Above $b^*(\varepsilon)=\tilde O(\tauc)$, blocks decode with a $B/b^*(\varepsilon)$-fold parallel speedup within the $\eps$ budget, while below it the within-block schedule is sequential. At a fixed budget of $T$ steps the within-block spacing is $BT/N$, so the bound applies once $B\ge (N/T)\,b^*(\eps)$. With the constants fitted in \cref{sec:empirics}, $b^*(\eps)$ exceeds practical block sizes, so likelihood comparisons across $B$ \citep{arriola2025} measure training error.

\subsection{Looped denoisers}\label{sec:looped}

LoopMDM \citep{loopmdm2026} loops a small shared block of early-middle layers in an MDM transformer and finds that moderate loop counts suffice for learning. We do not analyze transformer computation. Instead we bound an idealized looped class under an added source assumption: on order-$r$ Markov data the true conditional is computed by iterating a shared width-$r$ kernel, so each loop extends the receptive field by $r$ positions at zero parameter cost. The looped class learns this shared order-$r$ kernel, a correctly specified parametric model: its polynomial rate comes from the order-$r$ structure, and $\Lambda$ controls only the approximation error. Under CM alone the exponent's order in \cref{thm:upper} is optimal (\cref{thm:minimax}).

\begin{corollary}[Looped denoisers]\label{cor:looped}
Let $\pi$ be a stationary order-$r$ Markov source with transition floor $q_{\min}$, $(\tauc,\Czero)$ its constants from \cref{prop:markov}, and $\mathcal G_{r,q,\Lambda}$ the looped class of \cref{def:looped}: a shared order-$r$ kernel with entries in $[q,1]$, $q\le q_{\min}$, looped $\Lambda$ times per side, output floor $p_{\min}$ as in \cref{thm:upper}. Take
\begin{equation*}
	\Lambda = \left\lceil\frac{\tau_c}{r}\log\frac{6C_0(\lmax + 1)}{\eps'}\right\rceil.
\end{equation*}
If $n\ge\frac{32\lmax^2}{\varepsilon'^2}\left[|\cV|^{r+1}\log\frac{48L_{\mathrm{cov}}}{\varepsilon'}+\log\frac2\delta\right]$, with $L_{\mathrm{cov}}\le4r/q^{r+1}$ (exact constant in \cref{lem:loopcov}), then for every admissible $\omega$ the ERM $\hat f$ over $\mathcal G_{r,q,\Lambda}$ trained under $\omega$ satisfies $\Prob(\mathcal E_\omega(\hat f)\le\varepsilon')\ge1-\delta$.
\end{corollary}
\emph{Proof sketch.} We apply the same decomposition from \cref{thm:upper} and split the excess into approximation and estimation errors. The class contains the looped predictor built from the source's own kernel; after $\Lambda$ loops it has used the visible tokens within $\Lambda r$ positions on each side, so its total variation with the true distribution is at most $\frac32\Czero e^{-\Lambda r/\tauc}$. Each pass of the recursion damps differences between messages so a small change of the kernel entries changes the output by a comparable amount regardless of the number of loops, and the estimation error depends only on the $|\cV|^{r+1}$ kernel entries (\cref{lem:loopcov}). Each loop thus multiplies the approximation-error bound by $e^{-r/\tauc}$ until $\Lambda r$ reaches $\tauc\log(1/\eps')$, after which further loops do not reduce the bound; with the dilated schedule, inference needs $T\Lambda=\tilde O\left(\tauc\log N\cdot(\tauc/r)\log(1/\eps')\right)$ loop iterations.%

\section{Experiments}\label{sec:empirics}

\begin{table}[t]
\centering\footnotesize
\caption{Theory-motivated checks and results. Single runs; setups and estimators in \cref{app:experiments}.}
\label{tab:results}
\begin{tabular}{@{}p{0.50\linewidth}p{0.46\linewidth}@{}}
\toprule
Prediction & Result \\
\midrule
\multicolumn{2}{@{}l}{\emph{Influence on MDLM's conditionals}}\\
mean influence decays exponentially (\cref{ass:cm}) & exp.\ fit $R^2$ $0.989$ vs.\ power law $0.966$ \\
worst-case influence stays large (\cref{ass:strong}) & $95$th pct.\ $0.72$ at $d=64$ \\
dilated factorization term $\hat I\approx0$ by $b=16$ & yes; random order $\ne0$ at $64$ \\
\midrule
\multicolumn{2}{@{}l}{\emph{Block size on a synthetic source (\cref{cor:blockar})}}\\
held-out excess small at every $B$ & $\leq 0.031$ nats \\
no degradation for $B\ge 4b^*(1)=40$ & holds \\
\midrule
\multicolumn{2}{@{}l}{\emph{Decoding steps on MDLM and LLaDA-8B (\cref{cor:endtoend,cor:randorder})}}\\
dilated saturates within $[8,32]$ & fwd.\ $24$ (released), $32$ (finetunes); gen.\ $9.7$; LLaDA: none by $128$ \\
dilated saturates before random & fwd.\ $24$ vs.\ $96$, $32$ vs.\ $64$; gen.\ $9.7$ vs.\ $44.7$ \\
saturation well below $b^*(1)\approx860$ & $9.7$--$32$ steps \\
\bottomrule
\end{tabular}
\end{table}

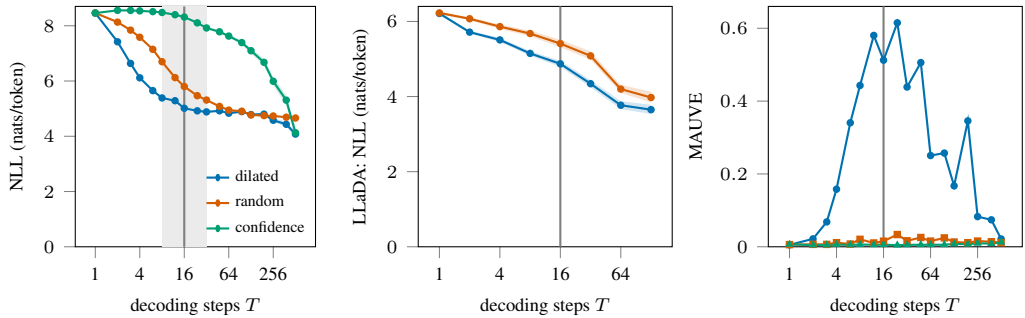
\begin{figure}[ht]
\centering
\providecolor{C0}{HTML}{0072B2}\providecolor{C1}{HTML}{D55E00}\providecolor{C2}{HTML}{009E73}%
\noindent
\begin{minipage}[t]{0.325\linewidth}\centering
\begin{tikzpicture}
\begin{axis}[width=0.70\linewidth,height=0.70\linewidth,scale only axis,xmode=log,log basis x=2,xtick={1,4,16,64,256},xticklabels={1,4,16,64,256},x tick label style={font=\fontsize{7}{8}\selectfont},y tick label style={font=\fontsize{7}{8}\selectfont},label style={font=\fontsize{7}{8}\selectfont},legend columns=1,legend style={font=\fontsize{6}{7}\selectfont,draw=none,fill=none,inner sep=1pt,at={(0.99,0.01)},anchor=south east},legend image post style={xscale=0.5},legend cell align=left,xlabel={decoding steps $T$},ylabel={NLL (nats/token)},ymin=0.0,ymax=8.7,tick align=outside,xtick pos=bottom,ytick pos=left]
\fill[black!8] (axis cs:8,0.0) rectangle (axis cs:32,8.7);
\draw[gray,thick] (axis cs:16,0.0) -- (axis cs:16,8.7);
\addplot[name path=dilL,draw=none,forget plot] table[x=b,y=dil_lo] {results/steps_quality.csv};
\addplot[name path=dilH,draw=none,forget plot] table[x=b,y=dil_hi] {results/steps_quality.csv};
\addplot[C0,opacity=0.18,forget plot] fill between[of=dilL and dilH];
\addplot[C0,mark=*,mark size=1pt,thick] table[x=b,y=dil_nll] {results/steps_quality.csv};
\addlegendentry{dilated}
\addplot[name path=rndL,draw=none,forget plot] table[x=b,y=rnd_lo] {results/steps_quality.csv};
\addplot[name path=rndH,draw=none,forget plot] table[x=b,y=rnd_hi] {results/steps_quality.csv};
\addplot[C1,opacity=0.18,forget plot] fill between[of=rndL and rndH];
\addplot[C1,mark=*,mark size=1pt,thick] table[x=b,y=rnd_nll] {results/steps_quality.csv};
\addlegendentry{random}
\addplot[name path=conL,draw=none,forget plot] table[x=b,y=con_lo] {results/steps_quality.csv};
\addplot[name path=conH,draw=none,forget plot] table[x=b,y=con_hi] {results/steps_quality.csv};
\addplot[C2,opacity=0.18,forget plot] fill between[of=conL and conH];
\addplot[C2,mark=*,mark size=1pt,thick] table[x=b,y=con_nll] {results/steps_quality.csv};
\addlegendentry{confidence}
\end{axis}\end{tikzpicture}
\end{minipage}%
\providecolor{C0}{HTML}{0072B2}\providecolor{C1}{HTML}{D55E00}\providecolor{C2}{HTML}{009E73}%
\begin{minipage}[t]{0.325\linewidth}\centering
\begin{tikzpicture}
\begin{axis}[width=0.74\linewidth,height=0.70\linewidth,scale only axis,xmode=log,log basis x=2,xtick={1,4,16,64},xticklabels={1,4,16,64},x tick label style={font=\fontsize{7}{8}\selectfont},y tick label style={font=\fontsize{7}{8}\selectfont},label style={font=\fontsize{7}{8}\selectfont},legend style={font=\fontsize{6}{7}\selectfont,draw=none,fill=none,at={(0.03,0.03)},anchor=south west},legend cell align=left,xlabel={decoding steps $T$},ylabel={LLaDA: NLL (nats/token)},ymin=0,ymax=6.4,tick align=outside,xtick pos=bottom,ytick pos=left]
\draw[gray,thick] (axis cs:16,0.0) -- (axis cs:16,6.4);
\addplot[name path=dilL,draw=none,forget plot] table[x=b,y=dil_lo] {results/llada.csv};
\addplot[name path=dilH,draw=none,forget plot] table[x=b,y=dil_hi] {results/llada.csv};
\addplot[C0,opacity=0.18,forget plot] fill between[of=dilL and dilH];
\addplot[C0,mark=*,mark size=1pt,thick] table[x=b,y=dil_nll] {results/llada.csv};
\addplot[name path=rndL,draw=none,forget plot] table[x=b,y=rnd_lo] {results/llada.csv};
\addplot[name path=rndH,draw=none,forget plot] table[x=b,y=rnd_hi] {results/llada.csv};
\addplot[C1,opacity=0.18,forget plot] fill between[of=rndL and rndH];
\addplot[C1,mark=*,mark size=1pt,thick] table[x=b,y=rnd_nll] {results/llada.csv};
\end{axis}\end{tikzpicture}
\end{minipage}%
\begin{minipage}[t]{0.325\linewidth}\centering
\begin{tikzpicture}
\begin{axis}[width=0.74\linewidth,height=0.70\linewidth,scale only axis,xmode=log,log basis x=2,xtick={1,4,16,64,256},xticklabels={1,4,16,64,256},x tick label style={font=\fontsize{7}{8}\selectfont},y tick label style={font=\fontsize{7}{8}\selectfont},label style={font=\fontsize{7}{8}\selectfont},legend style={font=\fontsize{6}{7}\selectfont,draw=none,fill=none,at={(0.97,0.97)},anchor=north east},legend cell align=left,xlabel={decoding steps $T$},ylabel={MAUVE},ymin=0.0,ymax=0.66,tick align=outside,xtick pos=bottom,ytick pos=left]
\draw[gray,thick] (axis cs:16,0.0) -- (axis cs:16,0.66);
\addplot[C0,mark=*,mark size=1pt,thick] table[x=b,y=dil_mauve] {results/mauve.csv};
\addplot[C1,mark=square*,mark size=1pt,thick] table[x=b,y=rnd_mauve] {results/mauve.csv};
\addplot[C2,mark=triangle*,mark size=1pt,thick] table[x=b,y=con_mauve] {results/mauve.csv};
\end{axis}\end{tikzpicture}
\end{minipage}%
\caption{Left: NLL under GPT-2-large (averaged over sequences) against decoding steps for dilated, random, and confidence-greedy schedules on MDLM; vertical line: the spacing at which the dilated $\hat I$ vanishes; gray band: half to twice that spacing; colored shading: bootstrap intervals. Middle: LLaDA-8B, dilated and random only, with MDLM's threshold. Right: MAUVE on MDLM ($512$ reference windows). Single runs; colors as on the left.}
\label{fig:steps}
\end{figure}

\Cref{tab:results} summarizes the first three experiments (details in \cref{app:experiments}). The influence measurements use a trained model's conditionals in place of the data's, so we treat these results as qualitative evidence.

\paragraph{Influence decays exponentially on average and stays large in the worst case.} On MDLM \citep{sahoo2024}, we replace the visible tokens beyond distance $d$ of a masked target with those of an unrelated held-out window and record the shift of the model's conditional in TV over random masking patterns. Mean influence decays exponentially, with decay scale ${\approx}26$ at short distances and $\hat\tau_c\approx81$ in the tail, and the exponential fits better than a power law (\cref{tab:influence}). The $95$th percentile is still $0.72$ at $d=64$. On this model's conditionals, the measurements are consistent with the averaged decay that \cref{thm:upper}(a) and \cref{cor:endtoend} use and rule out \cref{ass:strong} with useful constants. The dilated factorization term $\hat I$ is indistinguishable from zero by spacing $16$, while the random-order estimate is still nonzero at the commit size of spacing $64$.

\paragraph{Block size on an exactly computable source.} Six block-autoregressive models, identical except for $B\in\{4,\dots,128\}$, train on an order-$3$ synthetic source with exact conditionals. Decoding uses $T=N/4$ steps, so the within-block spacing is $B/4$, and \cref{cor:blockar}(ii) bounds the factorization error by $1$ nat once $B\ge4b^*(1)=40$, with $b^*(1)=10$ computed from influence constants fitted on the source (\cref{app:blocksize}). Held-out excess is at most $0.031$ nats at every $B$, and generation KL shows a factorization penalty only at $B=4$ and none above the threshold.

\paragraph{Dilated decoding saturates earlier than random order.} We score held-out text by each schedule's teacher-forced cross-entropy (forward sampling error; for random order, one fixed order per window, as in \cref{cor:randorder}) and samples by NLL under GPT-2-large \citep{radford2019gpt2} and MAUVE \citep{pillutla2021mauve} (\cref{fig:steps}). By \eqref{eq:tdelta}, dilated decoding saturates at $T=24$ on the released model against $96$ for random order ($32$ against $64$ after dilated or uniform finetuning), and it lies below random order at every matched step count $T\ge2$. The generation-NLL fit saturates at $9.7$ steps (bootstrap interval $[7.9,18.2]$) for dilated against $44.7$ for random. Dilated decoding thus needs two to five times fewer steps than random order, and every dilated saturation point lies within twice the spacing $16$ at which $\hat I$ vanishes, matching the ordering of \cref{cor:endtoend}(ii) and \cref{cor:randorder}. At the fitted constants, $b^*(1)\approx860$ is $27$--$89\times$ larger than the observed points, so the theorem's constants are loose while its predicted ordering holds. LLaDA-8B \citep{nie2025} is still improving at $128$ steps under both orders. Past ${\sim}64$ steps, MDLM's generation NLL keeps falling while MAUVE drops from its peak of $0.61$ at $T=24$, which \citet{zheng2024} attribute to reduced effective temperature in fp32 categorical sampling, the precision we use.

\paragraph{Loop count.} In a looped variant of MDLM trained from scratch, the loop count at which held-out loss saturates moves right with training, $2.42\to2.53\to2.62$ at $1$B, $2$B and $4$B tokens, the direction \cref{cor:looped} suggests. This experiment is qualitative and incomplete (\cref{app:loop}).

\section{Discussion}\label{sec:discussion}

\par We present an end-to-end KL bound for masked diffusion with a learned denoiser under conditional mixing. The sample budget is the same for every decoding schedule and minimax-optimal in its exponent's order, and we specialize the bound to dilated and block-autoregressive decoding and to a looped message-passing class on order-$r$ sources. Dilated decoding reaches KL $\eps+N\eps'$ in $\O(\tauc\log(N|\cV|/\eps))$ steps, where a fixed random order needs $\Omega(N)$ on an explicit source, and released and finetuned models show the same ordering. The guarantee holds for a denoiser trained under the decoding weighting, and its training term enters per token, so the sequence-level budget grows polynomially in $N$. Text with power-law pairwise mutual information lies outside conditional mixing, and for such sources dilated decoding needs polynomially many steps (\cref{rem:powerlaw}). Future work includes bounds for slower-mixing sources, schedules that avoid the context-free first step, tighter constants, influence probes on larger MDMs, and samplers that revise decoded tokens \citep{liu2024think,zhao2024correctors,wang2025remask}.

\begin{ack}
No third-party funding was received for this work; the A100 GPU time used in the experiments was rented. No competing interests are declared.
\end{ack}

\bibliographystyle{abbrvnat}
\bibliography{references}

\appendix
\begin{center}{\Large\bfseries Appendix}\end{center}
 
\section{Proofs}\label{app:verify}

\par We prove the two theorems presented in the paper with the help of a few observations about how distant tokens affect Markov-structured data and motivates our conditional mixing assumption as well as how much information we lose by ignoring certain distant tokens.
\par We structure Appendix~A by first proving in \cref{app:prelim} the lemmas and bounds that we use in our theorem proofs, analyzing in \cref{app:certificates} conditional mixing for finite-order and hidden Markov sources, connecting the sampler's learning error to excess risk in \cref{app:bridge}, and finally proving our main learning and decoding results in \cref{app:learning,app:orderings}.
 
\subsection{Bounding lemmas \& insights}\label{app:prelim}
 
\par We prove two lemmas used in later proofs and discuss insights they provide into how Markov sources and conditional mixing relate:
\begin{itemize}
	\item \cref{lem:forgetting} shows that products of positive matrices (and thus the posteriors of Markov-type sources) forget distant inputs exponentially fast
	\item \cref{lem:estimation} bounds the estimation error of ERM
\end{itemize}

\par We first briefly consider how much revealing a subset of tokens refines a token posterior. Given $x_C=\xi$ with $\pi(x_C=\xi)>0$ and disjoint $C,A\subseteq[N]\setminus\{i\}$, the posterior $\bar p=\pi(x_i\in\cdot\mid x_C=\xi)$ is the mixture $\E_\zeta[p_\zeta]$ of the refined posteriors $p_\zeta=\pi(x_i\in\cdot\mid x_C=\xi,\,x_A=\zeta)$ over $\zeta\sim\pi(x_A\in\cdot\mid x_C=\xi)$. Every measure of influence we present describes how spread out these components are.

We start by recalling that the worst-case coefficient $\kpi$ bounds the distance between two components; by Jensen's and convexity of total variation,
\begin{equation}\label{eq:mixtv}
	\TV(p_\zeta,\bar p)\le\E_{\zeta'}\TV(p_\zeta,p_{\zeta'})\le\kpi(d)\qquad\text{whenever }A\cap\cB(i,d)=\emptyset
\end{equation}

We also observe that strong conditional mixing implies conditional mixing $\bar\kappa \leq \kpi$ with the same constants: with $C=z\setminus A$ and $A\subseteq z\setminus\cB(i,d)$, \eqref{eq:mixtv} bounds $\TV(\pi(x_i\in\cdot\mid x_z),\pi(x_i\in\cdot\mid x_{z\setminus A}))$ by $\kpi(d)$ pointwise. Consequently, truncating the visible context to within $w$ tokens ($A = z\setminus \mathcal{B}(i,w)$) changes the posterior pointwise by at most $\kappa(w)$ and by \eqref{eq:kappaavg} likewise at most $\bar\kappa(w)$ in expectation under any masking distribution.

While influence is controlled in total variation, the sampler's error is an entropy and the learning error is a KL (excess risk). Consequently, we use Audenaert-Fannes to connect between entropy and total variation, allowing us to convert between all three: for distributions $p,q$ on $\cV$ with $t=\TV(p,q)$ and $h(t) \leq t\log{e/t}$ the binary entropy, the bound $|H(p)-H(q)|\le t\log(|\cV|-1)+h(t)$ of \citet{audenaert2007},
\begin{equation}\label{eq:etaenv}
	|H(p)-H(q)|\le\etamod\left(\TV(p,q)\right),\qquad\etamod(x)=x\log(e|\cV|/x), \etamod(0)=0
\end{equation}
where $\etamod$ is nondecreasing and concave on $[0,1]$.
 
\par For Markov-type sources the classical forward-backward recursion writes the posterior of a masked token as $p\propto R(\alpha\odot D_0\beta)$ for some matrix $R$ with messages $\alpha,\beta$ that are products of positive transfer matrices and of diagonal matrices that encode the visible tokens. Sequences that agree near the token share the nearest factors, so bounding $\kpi$ in \eqref{eq:mixtv} reduces to bounding how far such messages differ; we bound this difference in \cref{lem:forgetting} through Hilbert's projective metric $\dH(u,v)=\log\max_j(u_j/v_j)+\log\max_j(v_j/u_j)$, which compares positive vectors up to scale and satisfies the contraction $\tanh(\pdiam(M)/4)<1$ where $\pdiam(M)=\sup_{u,v\ge0}\dH(Mu,Mv)$ is its projective diameter \citep{birkhoff1957}.
 
\begin{lemma}[Exponential forgetting]\label{lem:forgetting}
Let $K_1,K_2,\dots,K_m$ be entrywise positive matrices with $\pdiam(K_j)\le\Delta$, $D_1,D_2,\dots,D_m$ be nonnegative diagonal matrices, and $\vartheta=\tanh(\Delta/4)$. Messages $\alpha=K_1D_1\cdots K_mD_mu$ and $\alpha'=K_1D_1\cdots K_mD_mu'$ that share these $m$ factors satisfy $\dH(\alpha,\alpha')\le\Delta\vartheta^{m-1}$ for all nonnegative $u,u'$. For entrywise positive $\alpha,\alpha',\beta,\beta'$, the posteriors $p\propto R(\alpha\odot D_0\beta)$ and $p'\propto R(\alpha'\odot D_0\beta')$ where $R$ is nonnegative and $D_0$ is nonnegative diagonal satisfy $p_j/p'_j\in[e^{-h},e^h]$ for every $j$ and $\TV(p,p')\le h$, where $h=\dH(\alpha,\alpha')+\dH(\beta,\beta')$.
\end{lemma}
 
\begin{proof}
We use standard facts about $\dH$ \citep{bushell1973,seneta2006}: it is defined for nonnegative vectors with the same support, with the maxima over that support, and is $\infty$ otherwise; it vanishes exactly when $u\propto v$ and is invariant under rescaling either argument. For entrywise positive $M$ the projective diameter is the maximal log cross-ratio,
\begin{equation}\label{eq:crossratio}
	\pdiam(M)=\max_{j,k,l,m}\log\frac{M_{jl}M_{km}}{M_{jm}M_{kl}},
\end{equation}
so $\pdiam(M)\le2\log(\max_{jk}M_{jk}/\min_{jk}M_{jk})$ and $\pdiam(M^\top)=\pdiam(M)$, and $\pdiam(MD)\le\pdiam(M)$ for nonnegative diagonal $D$, since the image of $MD$ lies in that of $M$. Birkhoff's theorem \citep{birkhoff1957} states that for entrywise positive $u,v$ with $Mu,Mv\ne0$,
\begin{equation}\label{eq:birkhoff}
	\dH(Mu,Mv)\le\tanh\left(\pdiam(M)/4\right)\,\dH(u,v).
\end{equation}
For the first claim, $K_mD_mu$ and $K_mD_mu'$ lie in the image of $K_mD_m$, whose diameter is at most $\Delta$, and each of the remaining $m-1$ factors $K_jD_j$ has diameter at most $\Delta$ and contracts $\dH$ by $\vartheta$ by \eqref{eq:birkhoff}. For the second claim, by triangle inequality $\dH(\alpha\odot D_0\beta,\alpha'\odot D_0\beta')\le h$. Nonnegative matrices preserve entrywise order and are homogeneous and thus do not increase $\dH$, so $\dH(p,p')\le h$ and the ratio bound follows. Then $\TV(p,p')=\sum_{j:\,p_j>p'_j}(p_j-p'_j)\leq 1-e^{-h}\le h$.
\end{proof}
 
\par \cref{lem:forgetting} bounds how much distant tokens move a posterior, which governs both the sampler's factorization error and the approximation error of a predictor that ignores distant tokens. To control the estimation error of ERM in the learning results, we next bound the estimation error of empirical risk minimization through the size of a finite cover of the predictor class; each training sequence contributes one bounded loss, so the bound depends only on the number of sequences $n$ however long each sequence is.
 
\begin{lemma}[Estimation over a finite cover]\label{lem:estimation}
Let $\mathcal G$ be a class of predictors whose outputs are entrywise $\geq p_{\min}$ with $\lmax=\log(1/p_{\min})$. Suppose that for some $\rho\in(0,\lmax]$ a finite set $\Phi_\rho$ of functions with values in $[-\lmax,0]$ satisfies $\min_{\varphi\in\Phi_\rho}\Vert\log f-\varphi\Vert_\infty\le\rho$ for every $f\in\mathcal G$. Let $\omega$ be admissible and let $\hat f$ minimize the empirical risk over $\mathcal G$ on $n$ independent draws from $\omega$. Then, w.p. $\geq 1-\delta$,
\begin{equation*}
	\mathcal E_\omega(\hat f)\le\inf_{f\in\mathcal G}\mathcal E_\omega(f)+2\lmax\sqrt{\frac{\log(2|\Phi_\rho|/\delta)}{2n}}+4\rho
\end{equation*}
If instead $\hat f$ minimizes the empirical masked cross-entropy \eqref{eq:ce}, the same holds at $\omega=\omega_{\mathrm u}$ with the last two terms multiplied by $(N+1)/N$.
\end{lemma}
 
\begin{proof}
	By Hoeffding and a union bound, with probability at least $1-\delta$ the empirical risk of every $\varphi\in\Phi_\rho$ (an average of $n$ i.i.d. losses in $[0,\lmax]$) is within $s_n=\lmax\sqrt{\log(2|\Phi_\rho|/\delta)/(2n)}$ of its expectation. Replacing a predictor by its nearest element in $\Phi_\rho$ changes both its empirical and its population risk by at most $\rho$, so for every $f\in\mathcal G$ passing from $\hat f$ to $f$ through their cover elements and the optimality of $\hat f$ yields $R_\omega(\hat f)\leq R_\omega(f)+2s_n+4\rho$. The masked cross-entropy follows by observing that each sequence's average loss over its masked positions is again an i.i.d. loss in $[0,\lmax]$ with mean $\frac{N}{N+1}R_{\omega_{\mathrm u}}$ by \eqref{eq:pretrainid}.
\end{proof}
 
\subsection{Influence decay in Markov sources}\label{app:certificates}
 
\par For Markov sources, visible tokens far from $x_i$ can only affect its posterior through the chain of positions between them; for transitions with a probability floor $q_{\min} > 0$ this influence thus contracts by a fixed factor, so the influence of visible tokens beyond distance $d$ decays geometrically with $d$. This directly motivates conditional mixing: strong CM (\cref{ass:strong}) assumes this exponential decay of influence itself and CM (\cref{ass:cm}) assumes it holds only on average over the source; this decay is what lets a width-$\O(\tauc\log(1/\eps'))$ window predictor approximate every conditional (\cref{thm:upper}) and makes tokens committed $b$ apart nearly independent conditioned on context (\cref{cor:endtoend}). We formalize this in \cref{prop:markov,prop:hmm} for finite-order Markov sources with a transition floor and for hidden Markov sources with a two-sided Doeblin kernel, with correlation length set by the floor. Both proofs apply \cref{lem:forgetting} to the forward-backward messages of the stationary process on $\bbZ$, whose influence coefficient bounds that of every length-$N$ marginal, so the constants do not depend on $N$.%
 
\begin{proof}[Proof of \cref{prop:markov}]
	The blocks $Y_t=x_{i+(t-1)r+1:i+tr}$, $t\in\bbZ$ form a stationary Markov chain with marginal $\mu_r$ and transition matrix $Q^{(r)}(y,y')=\prod_{s=1}^rQ(y'_s\mid y_{s:r},y'_{1:s-1})\in[q_{\min}^r,1]$; by \eqref{eq:crossratio}, $\pdiam(Q^{(r)})=\pdiam(Q^{(r)\top})\le\Delta_r$. Conditioning on $x_z=\xi$ multiplies block $t$ by the diagonal indicator $D_t$ of agreement with $\xi$, so the posterior of $x_i$ is the normalized last-coordinate marginal of $\alpha\odot D_0\beta$ with $\alpha=Q^{(r)\top}D_{-1}\cdots Q^{(r)\top}D_{-t}\mu_r$ and $\beta=Q^{(r)}D_1\cdots Q^{(r)}D_t\mathbf 1$ for any $t\ge m = \lfloor d/r\rfloor$ with $z\subseteq[i-tr,i+tr]$. Configurations equal on $z\cap\cB(i,d)$ give the same $D_s$ for $-(m-1)\le s\le m$, so $\beta$ and $\alpha$ share their first $m$ and $m-1$ factors, respectively, and for $m\geq 2$ \cref{lem:forgetting} gives
\begin{equation}\label{eq:markovkappa}
	\kpi(d)\le\Delta_r\vartheta_r^{m-1}+\Delta_r\vartheta_r^{m-2}\le2\Delta_r\vartheta_r^{\lfloor d/r\rfloor-2}\le2\Delta_r\vartheta_r^{d/r-3}=\Czero e^{-d/\tauc}.
\end{equation}
For $d<2r$ the right side exceeds $2\Delta_r>1\ge\kpi(d)$ since $q_{\min}\le\frac12$. Finally, $\log(1/\vartheta_r)=2\arctanh(q_{\min}^r)\ge2q_{\min}^r$ gives $\tauc\le r/(2q_{\min}^r)$.
\end{proof}
 
\begin{proof}[Proof of \cref{prop:hmm}]
By \eqref{eq:AZ}, the chain stays in $\supp\nu_{\mathrm D}$; there, $\Gamma$ is entrywise positive with cross-ratios at most $\eD^{-4}$, so $\pdiam(\Gamma)=\pdiam(\Gamma^\top)\le\Delta_\Gamma$ by \eqref{eq:crossratio}. With stationary distribution $\mu$ and emission matrices $G_s=\mathrm{diag}(g(\xi_{i+s}\mid\cdot))$ for visible positions $i+s\in z$ and $G_s=I$ otherwise, the posterior of $x_i$ is proportional to $R(\alpha\odot\beta)$ where $R_{ah}=g(a\mid h)$, $\alpha=\Gamma^\top G_{-1}\cdots\Gamma^\top G_{-t}\mu$, and $\beta=\Gamma G_1\cdots\Gamma G_t\mathbf 1$ for any $t\ge d$ with $z\subseteq[i-t,i+t]$; these messages are entrywise positive when $\pi(x_z=\xi)>0$. Configurations that agree on $z\cap\cB(i,d)$ yield messages that share their first $d$ factors on each side so for $d\ge1$, \cref{lem:forgetting} gives $\kpi(d)\le2\Delta_\Gamma\vartheta_\Gamma^{d-1}=\Czero e^{-d/\tauc}$ while $\kpi(0)\le1\le\Czero$ since $\Czero=8\log(1/\eD)/\tanh(\log(1/\eD))\ge8$. Thus, $\log(1/\vartheta_\Gamma)=2\arctanh(\eD^2)\ge2\eD^2$ gives $\tauc\le1/(2\eD^2)$.
\end{proof}
 
\subsection{From decoding to denoising}\label{app:bridge}
 
\par A decoder only queries the denoiser at the pairs $(i,z)$ at which it commits tokens. We show that this pair distribution is admissible even for adaptive decoders and that the learning term $\Elearn$ is $N$ times the denoiser's excess risk under this distribution. Thus, learning for a decoder is learning at its induced weighting, which is why \cref{thm:upper} is stated for every masking distribution and under \cref{ass:strong} for every admissible weighting.
 
\begin{proposition}[Adaptive strategies are admissible]\label{prop:adaptive}
	Let a decoding strategy $\nu$ choose at each step $k\geq 1$ a nonempty commit set $S_k\subseteq[N]\setminus z_k$, where the visible set is $z_k=S_1\cup\dots\cup S_{k-1}$, as a function of $(z_k,x_{z_k})$ and of a random variable $U\indep x$. Run $\nu$ on $x\sim\pi$ under teacher forcing so that committed positions take their true values, draw $i\sim\Unif([N])$ independently of $(x,U)$, and let $z=z_k$ for $k$ with $i\in S_k$. Then, the joint law $\omega_\nu$ of $(i,z,x)$ is admissible, and for every predictor $f$ the learning term of \eqref{eq:lz-decomp} \citep[Prop.~2]{lavenant2025} along the teacher-forced run of $\nu$ equals $\Elearn(f)=N\mathcal E_{\omega_\nu}(f)$; if $\nu$ is \emph{non-adaptive} (its selections depend on $U$ alone), then $\omega_\nu$ is a masking distribution.
\end{proposition}
 
\begin{proof}
A decoder decides to commit $i$ when $z$ is visible using only $x_z$ and randomness independent of $x$, so conditioning on that decision reveals nothing about $x_i$ beyond $x_z$. Visible sets only grow, so for $j\notin Z$ the event $\{i=j,\,z=Z\}$ is decided by $i$ and the selections made at visible sets inside $Z$, hence by $(i,x_Z,U)$. As $(i,U)\indep x$, the event is conditionally independent of $x_j$ given $x_Z$. For $\nu$ fixed before training, the per-sequence losses stay i.i.d.\ in $[0,\lmax]$, so \cref{lem:estimation} applies unchanged. By \eqref{eq:excessKL}, the learning term, the sum of the excess KLs over the $N$ commits, is $N$ times an average over a uniformly chosen position:
\begin{equation*}
	\Elearn(f)=\E\left[\sum_{k\ge1}\sum_{j\in S_k}\log\frac{\pi(x_j\mid x_{z_k})}{f(x_j\mid j,z_k,x_{z_k})}\right]=N\E_{\omega_\nu}\left[\log\frac{\pi(x_i\mid x_z)}{f(x_i\mid i,z,x_z)}\right]=N\mathcal E_{\omega_\nu}(f).
\end{equation*}
\end{proof}
 
\par Under $\alpha_t=1-t$, $\Prob(M=S)=\int_0^1t^{|S|}(1-t)^{N-|S|}\,\mathrm dt=\left((N+1)\tbinom{N}{|S|}\right)^{-1}$, so $M$ is uniform given its size, which is uniform on $\{0,\dots,N\}$. Thus, $\Prob(M\ne\emptyset)=N/(N+1)$, and given $M\ne\emptyset$ a uniform $i\in M$ with $z=[N]\setminus M$ has law $\omega_{\mathrm u}$, which is \eqref{eq:pretrainid}. The NELBO, \eqref{eq:ce} with $1/|M|$ replaced by $1/t$, equals $NR_{\omega_{\mathrm u}}(f)$: $1/t$ cancels the probability $t$ that a position $i$ is masked, and the same integral over the other $N-1$ positions gives $z$ its law given $i$ under $\omega_{\mathrm u}$.%
 
\subsection{Learning the denoiser}\label{app:learning}

\par By \cref{prop:adaptive}, it suffices to learn the true conditionals at a fixed weighting $\omega$. Under conditional mixing, visible tokens beyond distance $w$ change the posterior by $\mathcal{O}(\exp({-w/\tauc}))$; we thus consider predictors that read only a window of width $w$ then bound their approximation error by the influence (\cref{lem:approx}) and their estimation error by a cover (\cref{lem:estimation}), proving \cref{thm:upper}. \Cref{thm:minimax} then shows that the linear exponent in the sample complexity is necessary under conditional mixing alone, and \cref{cor:floored,cor:looped} improve it under a floor or known parametric structure.

\begin{definition}[Window class]\label{def:window}
For $w\geq 0$, the window context $\mathrm{ctx}_w(i,z,x)\in\cVm^{2w}$ lists $x_{i+o}$ for $o\in\{\pm1,\dots,\pm w\}$, with $\mask$ at masked or out-of-range positions. The window class is
\begin{equation*}
	\mathcal F_w=\left\{(1-\lambda)f+\lambda\Unif(\cV):\ f:\cVm^{2w}\to\mathcal P(\cV)\right\},\qquad\lambda=|\cV|p_{\min},
\end{equation*}
evaluated at the window context; its outputs are at least $p_{\min}$, so its log-losses lie in $[0,\lmax]$.
\end{definition}

\par By stationarity, the window conditional $\pi_w(\cdot\mid c)=\pi(x_0\in\cdot\mid x_o=c_o\forall o\text{ s.t. }c_o\ne\mask)$ satisfies $\pi(x_i\in\cdot\mid x_{z\cap\cB(i,w)})=\pi_w(\cdot\mid\mathrm{ctx}_w(i,z,x))$ for every $N$, $i\in[N]$, and $z\subseteq[N]\setminus\{i\}$.

\begin{lemma}[Approximation by window truncation]\label{lem:approx}
For every admissible $\omega$ and every $w\ge0$,
\begin{equation*}
	\min_{f\in\mathcal F_w}\mathcal E_\omega(f)\le(\lmax+1)\left(\bar\kappa_\omega(w)+\lambda\right),
\end{equation*}
where $\bar\kappa_\omega(w)=\E_\omega\left[\TV\left(\pi(x_i\in\cdot\mid x_z),\pi(x_i\in\cdot\mid x_{z\cap\cB(i,w)})\right)\right]$ is the averaged truncation error.
\end{lemma}

\begin{proof}
Take $f^\star=(1-\lambda)\pi_w+\lambda\Unif(\cV)\in\mathcal F_w$, which is within $\bar\kappa_\omega(w)+\lambda$ of the true conditional in expected total variation. For $\min_aq_a\ge p_{\min}$, using $\log(p_a/q_a)\le\lmax$ and $\log s\le s-1$,
\begin{equation*}
	\KL(p\Vert q)\le\sum_{a:\,p_a>q_a}\left[(p_a-q_a)\log\frac{p_a}{q_a}+q_a\log\frac{p_a}{q_a}\right]\le(\lmax+1)\TV(p,q),
\end{equation*}
and \eqref{eq:excessKL} gives $\mathcal E_\omega(f^\star)\le(\lmax+1)(\bar\kappa_\omega(w)+\lambda)$.
\end{proof}

\begin{proof}[Proof of \cref{thm:upper}]
Fix $w\ge0$, $p_{\min}\le1/(2|\cV|)$ and $\rho\in(0,\lmax]$. Rounding the $|\cVm|^{2w}|\cV|$ log-probabilities $\log f(a\mid c)$ where $c\in\cVm^{2w}, a\in\cV$ of each $f\in\mathcal F_w$ to the grid $\rho\bbZ\cap[-\lmax,0]$ yields a cover of size $\mathcal N_\rho$ with $\log\mathcal N_\rho\le|\cVm|^{2w}|\cV|\log(3\lmax/\rho)$, so \cref{lem:approx,lem:estimation} give w.p. $\geq 1-\delta$ the oracle inequality
\begin{equation}\label{eq:oracle}
	\mathcal E_\omega(\hat f)\le\underbrace{(\lmax+1)\left(\bar\kappa_\omega(w)+\lambda\right)}_{\text{approximation}}+\underbrace{2\lmax\sqrt{\frac{\log(2\mathcal N_\rho/\delta)}{2n}}+4\rho}_{\text{estimation}}.
\end{equation}
	We next show that the ERM under $\omega$ and the masked cross-entropy minimizer (for which $\omega=\omega_{\mathrm u}$) both satisfy $\mathcal E_\omega(\hat f)\le\eps'$ with probability $\geq 1-\delta$ whenever
\begin{equation}\label{eq:nexplicit}
	n\ge\frac{72\lmax^2}{\eps'^2}\left[|\cVm|^{2w}|\cV|\log\frac{72\lmax}{\eps'}+\log\frac2\delta\right],
\end{equation}
which gives the stated rate after substituting $w\le\tauc\log(4\Czero(\lmax+1)/\eps')+1$ into $|\cVm|^{2w}$.

\par\emph{Approximation.} With $s_0=\log(10|\cV|/\eps')$ and $p_{\min}=\eps'/(10|\cV|s_0)\leq 1/(2|\cV|)$, $\lmax+1=s_0+\log s_0+1\le2s_0$, so $(\lmax+1)\lambda\le\eps'/5$. By \eqref{eq:kappaavg} and \eqref{eq:mixtv}, $\bar\kappa_\omega(w)\le\bar\kappa(w)$ for masking distributions and $\bar\kappa_\omega(w)\le\kpi(w)$ for admissible $\omega$, so under \cref{ass:cm} in part (a) and \cref{ass:strong} in part (b), $w=\lceil\tauc\log(4\Czero(\lmax+1)/\eps')\rceil$ gives $(\lmax+1)\bar\kappa_\omega(w)\le\eps'/4$ and an approximation term below $\eps'/2$.

\par\emph{Estimation.} For the ERM under $\omega$, $\rho=\eps'/16$ and \eqref{eq:nexplicit} bound the square-root term by $\eps'/4$, so $\mathcal E_\omega(\hat f)\le\eps'/2+\eps'/4+\eps'/4=\eps'$. For the masked cross-entropy, $\rho=\eps'/24$ bounds the estimation term by $\eps'/3$, and the factor $(N+1)/N\le3/2$ of \cref{lem:estimation} gives $\mathcal E_{\omega_{\mathrm u}}(\hat f)\le\eps'/2+\eps'/2=\eps'$.
\end{proof}

\par The exponent of \cref{thm:upper} grows linearly in $\tauc\log|\cV|$ because the window class has $|\cVm|^{2w}$ contexts and $w\asymp\tauc\log(1/\eps')$; we next show that no learner avoids this growth:

\begin{proof}[Proof of \cref{thm:minimax}]
Each context of the $d-1$ most recent tokens hides one bit that sets the sign of a perturbation of size $\gamma\asymp\sqrt{\eps'}$; the perturbation is small enough to preserve strong conditional mixing, and a bit is observed only when its context occurs, so Assouad's lemma forces about $|\cV|^{d-1}/\eps'$ token observations.

\par\emph{Construction.} Let $m=\max\{1,(\tauc/2)\log(\Czero/2)\}$, $\eps_0=\min\{2^{-10},(\Czero e^{-3m/\tauc}/(160m))^2\}$ and $\gamma=8\sqrt{\eps'}$, so that $d$ is the largest integer with $20d\gamma e^{3d/\tauc}\le\Czero$ and $d\ge\lfloor m\rfloor$. Let $\chi,\psi:\cV\to\{-1,0,1\}$ sum to zero and vanish at no more than one symbol each, and let $\Pi=\{\pi_\theta:\theta\in\{\pm1\}^{\cV^{d-1}}\}$, where $\pi_\theta$ is the stationary order-$d$ chain with
\begin{equation*}
	\pi_\theta\left(x_t=a\mid x_{t-d}=\xi_1,\,x_{t-d+1:t-1}=y\right)=\frac{1+\gamma\theta_y\psi(\xi_1)\chi(a)}{|\cV|},\qquad\xi_1\in\cV, y\in\cV^{d-1}.
\end{equation*}

\par\emph{Strong conditional mixing.} As $t\gamma e^{3t/\tauc}\le\Czero/20$ at $t=d$ and at $t=m$ and $e^{2m/\tauc}\ge\Czero/2$, $d\gamma e^{d/\tauc}\le1/10$, so $\gamma\le\min\{1/(5d),1/4\}$ and $\arctanh\gamma\le1.1\gamma$. Whatever the visible tokens, $x_i$ enters the joint law through at most $d+1$ factors in $[1-\gamma,1+\gamma]/|\cV|$, so any two posteriors of $x_i$ satisfy
\begin{equation*}
	\TV\le\sinh\left(2(d+1)\arctanh\gamma\right)\le9d\gamma\le\Czero e^{-3d/\tauc},
\end{equation*}
which gives $\kpi(j)\le\Czero e^{-j/\tauc}$ for $j<3d$. For $j\ge3d$, the $d$-block kernel of the proof of \cref{prop:markov} has entries in $[(1-\gamma)^d,(1+\gamma)^d]/|\cV|^d$, hence diameter $\Delta\le4d\arctanh\gamma\le5d\gamma$ by \eqref{eq:crossratio} and $\bar\vartheta=\tanh(\Delta/4)\le1.25d\gamma\le e^{-d/\tauc}$, and \eqref{eq:markovkappa} with $(d,\Delta,\bar\vartheta)$ in place of $(r,\Delta_r,\vartheta_r)$ gives $\kpi(j)\le10d\gamma\,\bar\vartheta^{\lfloor j/d\rfloor-2}\le\Czero e^{-j/\tauc}$.

\par\emph{Bounding the excess.} As $\psi$ sums to zero, the uniform law on $\cV^d$ is stationary, so at $\omega^*$ the context $(\xi_1,y)$ is uniform with conditional $p^\pm_{\xi_1}=(1\pm\gamma\psi(\xi_1)\chi)/|\cV|$ for $\theta_y=\pm1$, and \eqref{eq:excessKL} separates the excess over the bits:
\begin{equation*}
	\mathcal E_{\omega^*}(f)=\sum_ye_y(f,\theta_y),\qquad e_y(f,\pm1)=\frac1{|\cV|^d}\sum_{\xi_1}\KL\left(p^\pm_{\xi_1}\Vert f(\cdot\mid\xi_1,y)\right).
\end{equation*}
For $\psi(\xi_1)\ne0$, $\TV(p^+_{\xi_1},p^-_{\xi_1})\ge\gamma/2$, so by Pinsker's inequality $\KL(p^+_{\xi_1}\Vert q)+\KL(p^-_{\xi_1}\Vert q)\ge\gamma^2/4$ for every $q$, and summing over the at least $|\cV|/2$ such $\xi_1$ gives $e_y(f,+1)+e_y(f,-1)\ge\delta=\gamma^2/(8|\cV|^{d-1})$. Flipping $\theta_y$ changes only the rows at contexts with suffix $y$, which each of the $N-d$ later tokens follows with probability $|\cV|^{-(d-1)}$, and $\KL(p^+_{\xi_1}\Vert p^-_{\xi_1})\le2\gamma\arctanh\gamma\le2.2\gamma^2$; so the two laws of the sample are within KL $2.2n(N-d)\gamma^2|\cV|^{-(d-1)}$, hence within total variation $\frac12$ by Pinsker's inequality when $n(N-d)<|\cV|^{d-1}/(300\eps')$, and Assouad's lemma \citep{yu1997} gives
\begin{equation*}
	\max_\theta\E\left[\mathcal E_{\omega^*}(\hat f)\right]\ge\frac12\sum_y\frac\delta2\left(1-\frac12\right)=\frac{|\cV|^{d-1}\delta}8=\frac{\gamma^2}{64}=\eps'.\qedhere
\end{equation*}
\end{proof}

\par The coefficient $2$ of $\tauc$ in \cref{thm:upper} comes from the linear conversion $\KL\le(\lmax+1)\TV$. When the source's own conditionals have a floor, the window conditional needs no smoothing, the conversion becomes quadratic, and the coefficient halves.

\begin{corollary}[Conditional floor]\label{cor:floored}
Let $\pi$ be stationary with $\pi(x_i=a\mid x_z=\xi)\ge q_c>0$ for all $i$, patterns $z\not\ni i$, $\xi$ with $\pi(x_z=\xi)>0$, and $a\in\cV$, and let $\pi$ satisfy \cref{ass:cm} in mean square:
\begin{equation*}
	\E_{x\sim\pi}\TV\left(\pi(x_i\in\cdot\mid x_z),\pi(x_i\in\cdot\mid x_{z\setminus A})\right)^2\le\Czero^2e^{-2d/\tauc}
\end{equation*}
for all $d\ge0$, $i$, $z\not\ni i$ and $A\subseteq z\setminus\cB(i,d)$. Then, for every masking distribution $\omega$ the proof of \cref{thm:upper} applies with $p_{\min}=\min\{q_c,(2|\cV|)^{-1}\}$, so $\lmax$ is constant in $\eps'$; with $w=\lceil\tauc\log(3\Czero/\sqrt{q_c\eps'})\rceil$ the sample complexity scales as $(1/\eps')^{2+\tauc\log|\cVm|+o(1)}$, so the $\tauc$-coefficient of the exponent halves from $2$ to $1$.
\end{corollary}

\begin{proof}
The window conditional satisfies $\pi_w\ge q_c\ge p_{\min}$ entrywise, so $\pi_w\in\mathcal F_w$, and for $\min_aq_a\ge q_c$,
\begin{equation*}
	\KL(p\Vert q)\le\log\left(1+\chi^2(p\Vert q)\right)\le\sum_a\frac{(p_a-q_a)^2}{q_a}\le\frac{4\TV(p,q)^2}{q_c}
\end{equation*}
As $(i,z)\indep x$, \eqref{eq:excessKL} and the mean-square condition with $A=z\setminus\cB(i,w)$ give $\mathcal E_\omega(\pi_w)\le4\Czero^2e^{-2w/\tauc}/q_c\le4\eps'/9<\eps'/2$. The rest of the proof of \cref{thm:upper} uses $p_{\min}$ only through $\lmax$, now constant, and substituting $w\le\tauc\log(3\Czero/\sqrt{q_c\eps'})+1$ into \eqref{eq:nexplicit} gives the rate.
\end{proof}

\par Mean-square conditional mixing is implied by \cref{ass:strong} through \eqref{eq:mixtv} and implies \cref{ass:cm} by Jensen's inequality. Order-$r$ sources have $q_c\ge q_{\min}^{r+1}$: $\pi(x_i=a\mid x_{-i})$ is a product of $r+1$ transition probabilities, each at least $q_{\min}$, over a probability at most $1$, and $\pi(x_i=a\mid x_z)$ averages it.

\par The window class learns a separate distribution $f(\cdot\mid c)$ for each of its $|\cVm|^{2w}$ contexts $c$. For an order-$r$ source, the forward--backward recursion of \cref{prop:markov} computes the true conditional from the $|\cV|^{r+1}$ entries of one kernel, each pass reaching $r$ positions further; learning the kernel and looping the recursion $\Lambda$ times gives a cover that does not grow with $\Lambda$ (\cref{lem:loopcov}) and an approximation error that decays in $\Lambda$, hence the polynomial rate of \cref{cor:looped}.

\begin{definition}[Looped class]\label{def:looped}
Let $q\in(0,1/|\cV|]$. For an order-$r$ kernel $Q_\theta$ with entries in $[q,1]$, the \emph{looped predictor} is $f_{\theta,\Lambda}=(1-\lambda)p_{\theta,\Lambda}+\lambda\Unif(\cV)$ with $\lambda=|\cV|p_{\min}$, where $p_{\theta,\Lambda}$ is the normalized last-coordinate marginal of $\alpha\odot D_0\beta$ for the messages $\alpha,\beta$ of the proof of \cref{prop:markov}, computed with the block kernel $Q^{(r)}_\theta$ by $\Lambda$ recursion steps each, from uniform initializations and with the visible tokens in blocks $-\Lambda,\dots,\Lambda$ entering as diagonal matrices. These predictors form the looped class $\mathcal G_{r,q,\Lambda}$.
\end{definition}

\begin{lemma}[Loop-count-free covering]\label{lem:loopcov}
Let $\Delta_r(q)=2r\log(1/q)$ and $\vartheta_r(q)=\tanh(\Delta_r(q)/4)$. Uniformly over $(i,z,\xi)$,
\begin{equation*}
	\left\Vert\log f_{\theta,\Lambda}-\log f_{\theta',\Lambda}\right\Vert_\infty\le L_{\mathrm{cov}}\Vert Q_\theta-Q_{\theta'}\Vert_\infty,\qquad L_{\mathrm{cov}}=\frac{4r}{q(1-\vartheta_r(q))}\le\frac{4r}{q^{r+1}}.
\end{equation*}
Hence, for $\rho\in(0,1]$, a grid of mesh $\rho/L_{\mathrm{cov}}$ on the $|\cV|^{r+1}$ kernel entries yields a cover of $\mathcal G_{r,q,\Lambda}$ at log-scale $\rho$, in the sense of \cref{lem:estimation}, with log-cardinality at most $|\cV|^{r+1}\log(3L_{\mathrm{cov}}/\rho)$.
\end{lemma}

\begin{proof}
Each entry of $Q^{(r)}_\theta$ is a product of $r$ entries of $Q_\theta$ in $[q,1]$, so $\Vert\log Q^{(r)}_\theta-\log Q^{(r)}_{\theta'}\Vert_\infty\le s=r\Vert Q_\theta-Q_{\theta'}\Vert_\infty/q$, and both kernels and their transposes have diameter at most $\Delta_r(q)$ by \eqref{eq:crossratio}. Write a message as $KD_1\cdots KD_\Lambda u$, and let $P_k$ use the kernel $K$ of $\theta$ in its first $k$ factors and the kernel $K'$ of $\theta'$ in the others. Then $P_ku$ and $P_{k-1}u$ are the images under $KD_1\cdots KD_{k-1}$ of $KD_kw$ and $K'D_kw$, which are within $2s$ in $\dH$, so by \eqref{eq:birkhoff}, whatever the visible tokens,
\begin{equation*}
	\dH(P_\Lambda u,P_0u)\le\sum_{k=1}^\Lambda2s\,\vartheta_r(q)^{k-1}\le\frac{2s}{1-\vartheta_r(q)}.
\end{equation*}
By \cref{lem:forgetting}, every ratio $p_{\theta,\Lambda}(a)/p_{\theta',\Lambda}(a)$ then lies in $[e^{-h},e^h]$ with $h\le4s/(1-\vartheta_r(q))=L_{\mathrm{cov}}\Vert Q_\theta-Q_{\theta'}\Vert_\infty$; mixing with $\lambda\Unif(\cV)$ only moves the ratios toward $1$, and $1-\vartheta_r(q)=2q^r/(1+q^r)\ge q^r$. One kernel from each of the at most $(3L_{\mathrm{cov}}/\rho)^{|\cV|^{r+1}}$ nonempty grid cells gives the cover.
\end{proof}

\begin{proof}[Proof of \cref{cor:looped}]
As $q\le q_{\min}$, the class contains $f_{Q,\Lambda}$ for the source's kernel $Q$. From the exact initializations, $\mathbf 1$ for $\beta$ and $\mu_r$ for $\alpha$, its loops would compute $\pi(x_i\in\cdot\mid x_{z\cap W})$ for the union $W\supseteq\cB(i,\Lambda r)$ of blocks $-\Lambda,\dots,\Lambda$; the uniform start is exact up to scale for $\beta$ and shares $\Lambda$ factors with the exact one for $\alpha$. By \eqref{eq:mixtv} and \cref{lem:forgetting}, using $\Czero=2\Delta_r/\vartheta_r^3$ and $\vartheta_r^\Lambda=e^{-\Lambda r/\tauc}$,
\begin{equation*}
	t_\Lambda=\sup_{(i,z,\xi)}\TV\left(p_{Q,\Lambda},\pi(x_i\in\cdot\mid x_z)\right)\le\kpi(\Lambda r)+\Delta_r\vartheta_r^{\Lambda-1}\le\frac32\Czero e^{-\Lambda r/\tauc}.
\end{equation*}
As in \cref{lem:approx}, $\min_{\mathcal G_{r,q,\Lambda}}\mathcal E_\omega\le(\lmax+1)(t_\Lambda+\lambda)$ for every admissible $\omega$, and the stated $\Lambda$ gives $(\lmax+1)t_\Lambda\le\eps'/4$; the proof of \cref{thm:upper} then applies with $t_\Lambda$ in place of $\bar\kappa_\omega(w)$ and the cover of \cref{lem:loopcov} at $\rho=\eps'/16$.
\end{proof}

\subsection{End-to-end guarantees}\label{app:orderings}

\par The sampler's KL splits into a learning term and the factorization error \eqref{eq:lz-decomp}. \Cref{prop:adaptive,thm:upper} bound the learning term by $N\eps'$ for a decoding strategy whose denoiser is trained under the strategy's weighting, which is \cref{cor:endtoend}(i). The factorization error is the total correlation of the tokens committed in each step, and under conditional mixing tokens far apart carry little information about each other given the context, by \eqref{eq:mixtv} and \eqref{eq:etaenv}; a schedule should therefore commit only distant tokens together, as the dilated schedule does, which gives \cref{cor:endtoend}(ii). A random order commits nearby tokens together and pays for each such pair, which gives the lower bound of \cref{cor:randorder}, and block-autoregressive decoding inherits both parts of \cref{cor:endtoend} block by block (\cref{cor:blockar}).

\begin{proof}[Proof of \cref{cor:endtoend}(i)]
By \eqref{eq:lz-decomp}, valid for adaptive strategies \citep[Prop.~2]{lavenant2025}, and \cref{prop:adaptive},
\begin{equation*}
	\KL(\pi\Vert p_{\mathrm{alg}})\le\Efact+N\,\mathcal E_{\omega_\nu}(\hat f)\le\Efact+N\eps'
\end{equation*}
with probability at least $1-\delta$, by \cref{thm:upper}(a) when $\nu$ is non-adaptive, so that $\omega_\nu$ is a masking distribution, and by \cref{thm:upper}(b) under \cref{ass:strong} otherwise.
\end{proof}

\begin{proof}[Proof of \cref{cor:endtoend}(ii)]
At step $t\in[b]$ the schedule commits the residue class $S_t=\{i_1<\dots<i_{|S_t|}\}$ of $t$ modulo $b$ given the visible set $z_t=S_1\cup\dots\cup S_{t-1}$, so by the chain rule, with $A=\{i_1,\dots,i_{m-1}\}$,
\begin{equation*}
	\Efact=\sum_{t=1}^b\TC_\pi(x_{S_t}\mid x_{z_t})=\sum_{t=1}^b\sum_{m=2}^{|S_t|}I(x_{i_m};x_A\mid x_{z_t}).
\end{equation*}
Each summand is the expected entropy change $\E[H(\bar p)-H(p_\zeta)]$, with $p_\zeta$ and $\bar p$ the posteriors of $x_{i_m}$ given $x_{z_t\cup A}$ and given $x_{z_t}$, so \eqref{eq:etaenv} and Jensen's inequality for the concave $\etamod$ give
\begin{equation}\label{eq:factbound}
	\Efact\le\sum_{t,m}\E\left[\etamod\left(\TV(p_\zeta,\bar p)\right)\right]\le(N-b)\etamod\left(\bar\kappa_\nu(b-1)\right)\le(N-b)\,\etamod\left(\bar\kappa(b-1)\right),
\end{equation}
where
\begin{equation}\label{eq:kappabar}
	\bar\kappa_\nu(b-1)=\frac1{N-b}\sum_{t=1}^b\sum_{m=2}^{|S_t|}\E\left[\TV(p_\zeta,\bar p)\right]\le\bar\kappa(b-1)
\end{equation}
is the influence averaged over the pairs committed together, bounded by \eqref{eq:kappaavg} with $z_t\cup A$ as the visible set; since the choice of $b$ below uses only that $\etamod$ is nondecreasing, \cref{cor:endtoend}(ii) and \cref{cor:blockar}(ii) also hold with $\bar\kappa_\nu(b-1)$ in place of $\bar\kappa(b-1)$. For $\eps\in(0,N\log|\cV|]$ (for larger $\eps$ every $b$ works), let
\begin{equation}\label{eq:bcal}
	L^*=\log(e|\cV|)+\log\left(1+\frac{2\Czero N}{\eps}\right),\qquad b=b^*(\eps)=1+\left\lceil\tauc\log\frac{2\Czero NL^*}{\eps}\right\rceil.
\end{equation}
Then $\bar\kappa(b-1)\le\Czero e^{-(b-1)/\tauc}\le s=\eps/(2NL^*)\le1$ and, as $\Czero\ge1$, $\log(e|\cV|)+\log\frac{2N}\eps\le L^*$, so
\begin{equation*}
	(N-b)\,\etamod\left(\bar\kappa(b-1)\right)\le Ns\log\frac{e|\cV|}{s}=\frac{\eps}{2L^*}\left(\log(e|\cV|)+\log\frac{2NL^*}{\eps}\right)\le\frac{\eps}{2L^*}\left(L^*+\log L^*\right)\le\eps.
\end{equation*}
If $b^*(\eps)\ge N$, take $b=N$: every step commits a single token, so $\Efact=0$. In both cases $\Efact\le\eps$ in $T=b$ steps, and part (i) applies to this non-adaptive schedule.
\end{proof}

\par The dilated schedule reaches factorization error $\eps$ in $\O(\tauc\log(N/\eps))$ steps. On the two-state chain $\pi_q$ a fixed random order needs $\Omega(N)$: its expected KL is a weighted sum of increments of the information profile \citep[Lems.~6 and~8]{lavenant2025}, and each increment is at least $c_q/N$, because a newly revealed token is the target's neighbor with probability $1/N$ and neighbors share a constant amount of information.

\begin{proof}[Proof of \cref{cor:randorder}]
Fix the step sizes, with cumulative counts $0=a_0<a_1<\dots<a_T=N$, gaps $g_k=a_k-a_{k-1}$ and $r_a(i)=\min\{a_k:a_k\ge i\}$. Let $F(j)=\E[I(x_t;x_{z_j})]$ be the information profile \citep[cf.][Sec.~4.1]{lavenant2025}, with $t\sim\Unif([N])$ and $z_j$ an independent uniform $j$-subset of $[N]\setminus\{t\}$, and let $\Delta F(i)=F(i)-F(i-1)$. For an order uniform given the sizes, \citet[Lems.~6 and~8]{lavenant2025} give
\begin{equation*}
	\E_\sigma\KL(\pi\Vert p^\sigma)=\sum_{i=1}^{N-1}\Delta F(i)\left(r_a(i)-i\right),
\end{equation*}
so it suffices to bound the increments from below.

\par\emph{Each increment is at least $c_q/N$.} Let $z'$ be a uniform $i$-subset of $[N]\setminus\{t\}$, $U\sim\Unif(z')$ and $z=z'\setminus\{U\}$. As $z$ is a uniform $(i-1)$-subset, the chain rule gives $\Delta F(i)=\E[I(x_t;x_U\mid x_z)]$, and the event $G=\{t\le N-1,\ U=t+1\}$ that the new token is the target's right neighbor has probability $\frac{N-1}N\cdot\frac i{N-1}\cdot\frac1i=\frac1N$. On $G$, fix $x_z=\xi$ and let $s_-<t$ and $s_+>t+1$ be the nearest positions of $z$, if any. By the Markov property,
\begin{equation*}
	P(u,v)=\Prob(x_t=u,x_{t+1}=v\mid x_z=\xi)=\varphi(u)Q(u,v)\psi(v)/Z,\qquad Q=\begin{psmallmatrix}1-q&q\\q&1-q\end{psmallmatrix},
\end{equation*}
with $\varphi=Q^{t-s_-}(\xi_{s_-},\cdot)$ and $\psi=Q^{s_+-t-1}(\cdot,\xi_{s_+})$, each replaced by $(\frac12,\frac12)$ if there is no such visible position, and $Z\le1$ as $\varphi,\psi$ are probability vectors. The entries of $Q^g$ are $\frac12(1\pm(1-2q)^g)\in[q,1-q]$ for $g\ge1$, so $\varphi_0\varphi_1,\psi_0\psi_1\ge q(1-q)$ and $\det P=\varphi_0\varphi_1\psi_0\psi_1\det Q/Z^2\ge q^2(1-q)^2(1-2q)$. As $P$ differs from the product of its marginals by $\det P\begin{psmallmatrix}1&-1\\-1&1\end{psmallmatrix}$, at total variation $2\det P$, Pinsker's inequality gives $I(x_t;x_{t+1}\mid x_z=\xi)\ge8(\det P)^2\ge c_q$, so $\Delta F(i)\ge\Prob(G)\,c_q=c_q/N$.

\par\emph{Summing over the steps.} As $r_a(i)-i$ runs through $0,\dots,g_k-1$ on the $k$-th gap and $\sum_kg_k=N$, Cauchy-Schwarz gives
\begin{equation*}
	\E_\sigma\KL(\pi\Vert p^\sigma)\ge\frac{c_q}N\sum_{k=1}^T\frac{g_k(g_k-1)}2\ge\frac{c_q}{2N}\left(\frac{N^2}T-N\right)=\frac{c_q}2\left(\frac NT-1\right).
\end{equation*}
Averaging over the step sizes and applying Jensen's inequality to the convex map $T\mapsto N/T$ gives $\eps\ge\frac{c_q}2\left(\frac N{\E[T]}-1\right)$, that is, $\E[T]\ge c_qN/(c_q+2\eps)$.
\end{proof}

\par Block-autoregressive decoding generates blocks left to right and runs Algorithm~1 inside each block; the chain rule splits its KL exactly over blocks, so part (i) of \cref{cor:endtoend} applies to the mixture of the within-block weightings and part (ii) applies inside each block.

\begin{proof}[Proof of \cref{cor:blockar}]
For (i), let $C_k$ be the $k$-th block, $p_k(\cdot\mid x_{C_{<k}})$ the law of the block that Algorithm~1 generates on it given the prefix, and $\omega_{\nu_k}$ the weighting of \cref{prop:adaptive} for its within-block strategy $\nu_k$. The induced weighting $\omega_B$ draws $J\sim\Unif([m])$ and $(i,z')\sim\omega_{\nu_J}$ with $z'\subseteq C_J$ and sets $z=C_{<J}\cup z'$; it is a masking distribution because each $\nu_k$ is $x$-independent. By the chain rule,
\begin{equation*}
	\KL(\pi\Vert p_{\mathrm{alg}})=\sum_{k=1}^m\E_{x_{C_{<k}}\sim\pi}\left[\KL\left(\pi(x_{C_k}\in\cdot\mid x_{C_{<k}})\,\big\Vert\,p_k(\cdot\mid x_{C_{<k}})\right)\right],
\end{equation*}
and by \eqref{eq:lz-decomp} and \cref{prop:adaptive} for block $k$'s run the $k$-th term is at most $\E_\pi[\Efact^{(k)}]+B\,\E_\pi[\mathcal E^{(k)}(\hat f)]$, with $\mathcal E^{(k)}(\hat f)$ the excess of $\hat f$ at $\omega_{\nu_k}$ given the prefix. By \eqref{eq:excessKL}, $B\sum_k\E_\pi[\mathcal E^{(k)}(\hat f)]=Bm\,\mathcal E_{\omega_B}(\hat f)=N\mathcal E_{\omega_B}(\hat f)$, which \cref{thm:upper}(a) bounds by $N\eps'$ with probability at least $1-\delta$.

\par For (ii), let $b=\min\{b^*(\eps),B\}$. Applying \eqref{eq:factbound} inside each block, with the prefix added to the visible set and the expectation taken jointly over it, block $k$ contributes $B-b$ summands, so $\Efact^{\mathrm{blk}}\le m(B-b)\,\etamod(\bar\kappa(b-1))\le N\etamod(\bar\kappa(b-1))$. If $b^*(\eps)\le B$, the chain following \eqref{eq:bcal} bounds this by $\eps$ with $T=mb=(N/B)\,b^*(\eps)$ steps; otherwise $b=B$, so $\Efact^{\mathrm{blk}}=0$ and $T=mB=N$.
\end{proof}

\section{Experimental details}\label{app:experiments}
 
\subsection{Setup}\label{app:setup}
 
\paragraph{Benchmarks.} All text experiments use the OpenWebText dataset with GPT-2 tokenizer. Training draws from the first 7.9M documents, and evaluation uses the last 100K documents. The block-size experiment (\cref{app:blocksize}) uses a synthetic order-$3$ Markov source over four symbols with exact conditionals, so held-out excess and generation KL are measured against the exact conditionals and entropy.
 
\paragraph{Models.} We finetune MDLM \citep{sahoo2024}, and train from scratch a looped variant of MDLM (\cref{app:loop}) and six small block-AR models on a synthetic source (\cref{app:blocksize}). \Cref{tab:training} lists the training settings; each experiment's subsection gives its architecture. All experiments ran on a single NVIDIA A100-SXM4-40GB; the full set of experiments took $\sim 130$ GPU-hours.
 
\begin{table}[h]
\caption{Training settings of the models trained in this paper.
Weight decay for the looped model applies to matrices only.
The dilated finetune and the uniform control share the same drawn
training windows and differ in the masking law and loss target
alone.}
\label{tab:training}
\centering\footnotesize\setlength{\tabcolsep}{3pt}
\begin{tabular}{@{}lp{2.35cm}p{2.35cm}p{2.45cm}p{2.45cm}@{}}
\toprule
& dilated finetune & uniform control & looped model & block models \\
\midrule
Initialization & released MDLM & released MDLM & scratch, seed $0$ & scratch, seed $0$, shared across $B$ \\
Data & OWT train, seed $0$ & same windows as finetune & OWT train & synthetic, $20{,}000$ seqs \\
Context & $1024$ & $1024$ & $1024$ & $256$ \\
Tokens / steps & $0.25$B / $7630$ & $0.25$B / $7630$ & $4$B / $7630$ & plateau rule, $1500$--$8000$ steps \\
Batch (sequences) & $32$ & $32$ & $512$ & $256$ \\
Optimizer & AdamW $(0.9,0.999)$ & AdamW $(0.9,0.999)$ & AdamW $(0.9,0.95)$ & AdamW $(0.9,0.999)$ \\
Learning rate & $10^{-5}$ & $10^{-5}$ & $3\times10^{-4}$ & $3\times10^{-4}$ \\
Warmup, then & $100$, constant & $100$, constant & $2500$, constant & $500$, constant \\
Weight decay & $0$ & $0$ & $0.1$ & $0.01$ \\
Gradient clip & $1.0$ & $1.0$ & $1.0$ & none \\
Dropout & $0.1$ & $0.1$ & $0.1$ & none \\
Precision & bf16, fp32 loss & bf16, fp32 loss & bf16, fp32 head & fp32 \\
Masking law & dilated, $b\in\{8,16,32,64\}$ & uniform $t\sim U(0,1)$ & uniform, loop $\Lambda\sim U\{1..12\}$ & block-$B$ prefix \\
Loss & unweighted CE, class $k$ & unweighted CE, all masked & ELBO $1/t$ & unweighted CE, block $k$ \\
\bottomrule
\end{tabular}
\end{table}
 
\paragraph{Metrics.} We measure 
\begin{itemize}
	\item \emph{Influence coefficient}: $\hat{\kappa}(d)$ is the TV between a model's conditional at a masked position under two contexts that only differ beyond distance $d$ of the position
	\item \emph{Forward sampling error}: the teacher-forced CE $\frac1N\E_\pi[-\log p_{\mathrm{alg}}(x)]=\frac1N\left(H(\pi)+\KL(\pi\Vert p_{\mathrm{alg}})\right)$ of a decoding schedule, whose KL term is the quantity we bound
	\item \emph{Generation quality}: the mean per-sequence NLL of generated text, MAUVE \citep{pillutla2021mauve} against held-out text, and masked NELBO
	\item \emph{Held-out excess and generation KL}: the excess risk and $\KL(\pi\Vert p_{\mathrm{alg}})$ on the synthetic source
\end{itemize}
 
\subsection{Influence measurement}\label{app:influence}
 
We measure how strongly visible tokens beyond distance $d$ move a model's conditional at a masked position, as an estimate of the influence coefficients of \cref{ass:cm,ass:strong}, and the dilated factorization term $\hat I$, which sets the threshold for the decoding experiments. We ask whether mean influence decays exponentially or as a power law, and whether the worst case decays.%
 
\paragraph{Setting.} We use the released MDLM checkpoint \citep{sahoo2024} (12 blocks, hidden size 768, 128-dimensional conditioning pathway, context 1024) over the GPT-2 vocabulary with a mask token. The checkpoint ignores its time input, so no time conditioning enters the measurement. Each measurement uses two distinct windows drawn uniformly from held-out OWT windows of length 1024. $\hat I$ is estimated on 192 windows of the eval stream.
 
\paragraph{Procedure.} We draw two sequences $x, y$. For a masked target position $i$ under a visibility pattern $z$, let $x'$ equal $x$ at every unmasked position within $d$ of $i$ and $y$ beyond. The influence at distance $d$ is the total variation between the model's two conditionals at $i$,
\begin{equation}\label{eq:kappahat}
\hat\kappa(d)= \TV\left(f(\cdot\mid x_z),f(\cdot\mid x'_z)\right)
\end{equation}
computed exactly from the token logits. We make the target $i$ uniform on $[64,961)$ so that the window around it lies within the context; visibility is i.i.d. Bernoulli($v$) for $v\in \{0.25,0.5,0.9\}$ and the target masked, and we estimate for distances $d\in\{1,2,4,8,16,24,32,48,64\}$. Each $(d,v)$ cell uses 1000 sampled configurations, and we record the mean and 95th percentile of \cref{eq:kappahat} per cell.
 
The decay scale $\hat\tau_c$ and prefactor $\widehat\Czero$ come from an exponential least-squares fit of the per-distance means on the widest contiguous window of distances.

For a dilated step at spacing $b$ with commit set $S$ and previously revealed classes $z$, we estimate the per-summand factorization term by $\hat I(b)$, the mean of
\begin{equation}\label{eq:ihat}
\log f\left(x_i\mid x_z,x_{S\setminus\{i\}}\right)-\log f\left(x_i\mid x_z\right)
\end{equation}
over the $192$ held-out windows and over tokens $i\in S$ committed after other tokens of the same step; it estimates the per-summand mutual information $I(x_i;x_{S\setminus\{i\}}\mid x_z)$ that the proof of \cref{cor:endtoend}(ii) bounds. The random-order value uses commit sets of the same size. Its interval is a $95\%$ percentile bootstrap interval from $1000$ resamples of the windows; ``indistinguishable from zero'' means the interval contains zero.
 
\begin{table}[h]
\caption{Influence measurement on MDLM's conditionals.}
\label{tab:influence}
\centering\small
\begin{tabular}{@{}lc@{}}
\toprule
$R^2$, mean decay exponential fit & $0.989$ \\
$R^2$, mean decay power-law fit & $0.966$ \\
$(\hat\tau_c,\widehat\Czero)$ on the tail window; interval for $\hat\tau_c$ & $(81.1,\,1)$; $[43,95]$ \\
local decay scale over the first distances & ${\approx}26$ \\
$95$th percentile of $\hat\kappa$ at $d=1$ / $d=64$ & $0.999$ / $0.722$ \\
$\hat I(16)$, dilated: point/upper interval edge (per summand) & $0.0036$ / $0.042$ \\
$\hat I(64)$, random order (per summand) & $0.087$ \\
\bottomrule
\end{tabular}
\end{table}
 
\paragraph{Results.} The exponential fits the mean decay better than the power law (\cref{tab:influence}), with decay scale $81.1$ on the tail window and interval $[43,95]$; the first few distances decay faster at a local scale near $26$. The $95$th percentile falls only from $0.999$ at $d=1$ to $0.722$ at $d=64$. The dilated $\hat I$ is indistinguishable from zero by spacing $16$ while the random-order interval still excludes zero at the commit size of spacing $64$. The spacing \eqref{eq:bcal} of \cref{cor:endtoend}(ii) at these estimates is $b^*(1)\approx 860$ at $N=1024$.
 
On the model's conditionals, averaged influence decays fast and the worst case does not; \cref{thm:upper}(a) and \cref{cor:endtoend} rely only on the averaged behavior. Against the observed dilated saturation points ($9.7$ to $32$ steps), $b^*(1)\approx860$ is larger by $27$--$89\times$. The exchange construction measures the released model's conditionals, which stand in for the data's conditionals that the theory concerns.
 
\subsection{Block size on an exactly computable source}\label{app:blocksize}
 
We train block-AR models at six block sizes on a synthetic source with exact conditionals, then measure held-out excess and generation KL. \Cref{cor:blockar}(ii) predicts generation KL flat in the block size $B$ above the threshold $4b^*(1)$.
 
\paragraph{Setting.} Each model is a four-layer pre-norm bidirectional transformer of width $128$, GELU feed-forward block of width $512$, and learned position embeddings, over a five-symbol vocabulary (four source symbols and a mask token) at context $256$. The six models differ only in $B\in\{4,8,16,32,64,128\}$ and have the same initialization.
 
The source is a stationary order-3 Markov chain over the four symbols whose conditional at each context is an equal mixture of uniform and $\mathrm{Dir}(2,2,2,2)$. This yields $q_{\min}=0.136$ and, over a restricted family of masks, influence $0.370,0.148,0.053,0.020$ for $j=1,\dots,4$ and below $10^{-7}$ by $j=16$, fit by $\tauc=0.977$, $C_0=2$, which gives $b^*(1)=10$ and the threshold $4b^*=40$.
 
\paragraph{Procedure.} We draw a block index $k$ uniformly on $\{0,\dots,N/B-1\}$, reveal blocks before $k$, mask block $k$ i.i.d. Bernoulli$(t)$ with $t$ uniform on $[0,1)$, and take the mean cross-entropy on the masked positions of block $k$. Decoding runs for $T=N/4$ steps at every $B$ with held-out excess measured against the true conditionals and generation KL the mean of $-\log p_{\mathrm{alg}}$ over 4096 sequences drawn from the source $\pi$, minus the exact window entropy.
 
\paragraph{Results.} Held-out excess is at most $0.031$ nats at every $B$; generation KL is flat at about $N\hat\eps'$, with $\hat\eps'$ the held-out excess, from $B=8$ up and shows a factorization penalty only at $B=4$; the observed saturation is near $B=7$, against the computed window $[20,80]$ of half to twice the threshold $40$.
 
The comparison isolates factorization, and the computed window is conservative by about $5\times$.
 
\subsection{Decoding steps}\label{app:decoding}
 
The factorization error of dilated decoding at spacing $b$ is at most $N-b$ times the per-summand mutual information that $\hat I(b)$ estimates (\cref{app:influence}), so \cref{cor:endtoend}(ii) predicts that dilated decoding saturates within $[8,32]$, half to twice the spacing $16$ at which $\hat I$ vanishes, and \cref{cor:randorder} predicts that random order saturates later. We test both predictions on the forward sampling error (\cref{app:forward}) and on generated samples (\cref{app:generation}).
 
\subsubsection{Forward sampling error}\label{app:forward}
 
We score text under each decoding schedule, measuring the forward sampling error $\frac1N\E_\pi[-\log p_{\mathrm{alg}}(x)]$ without generating samples. For random order we score one fixed random order per window, the per-order error of \cref{cor:randorder}; the order-averaged generator has lower cross-entropy by convexity.
 
\paragraph{Setting.} We score MDLM and two finetunes of MDLM on 0.25B tokens from OWT (7630 steps of 32 sequences), one with dilated masking and the other with standard uniform masking, using the same data windows as in \cref{app:generation} except all table cells share the same 512 windows.
 
\paragraph{Procedure.} Let $S_1,\cdots,S_T$ be a schedule's commit sets and $z_k=S_1\cup\dots\cup S_{k-1}$ the positions visible before step $k$. Then, the score of a window after $T$ steps is 
\begin{equation}\label{eq:forwardce}
\mathrm{CE}(T) = -\frac1N\sum_{k=1}^{T}\sum_{i\in S_k} \log f\left(x_i\mid x_{z_k}\right)
\end{equation}
where $f$ is the model's masked conditional. We score for $T\in\{1,2,4,8,16,24,32,48,64,96,128\}$ with 95\% bootstrap intervals over sequences from 1000 resamples. We calculate saturation as
\begin{equation}\label{eq:tdelta}
T_\delta = \min\left\{T: \mathrm{CE}(T)-\mathrm{CE}(128)\le\delta\right\}
\end{equation}
where the tolerance $\delta=0.03$ is the median half-width of the cells' intervals so that $T_\delta$ is the first count at which a curve is within resolution of its value at the largest count.
 
\begin{table}[h]
\caption{Forward sampling error at all step counts for the six model-schedule pairs.}
\label{tab:forward}
\centering\footnotesize\setlength{\tabcolsep}{3.5pt}
\begin{tabular}{@{}lccccccccccc@{}}
\toprule
$T$ & $1$ & $2$ & $4$ & $8$ & $16$ & $24$ & $32$ & $48$ & $64$ & $96$ & $128$ \\
\midrule
released, dilated & $7.556$ & $4.654$ & $3.703$ & $3.381$ & $3.260$ & $3.241$ & $3.230$ & $3.235$ & $3.219$ & $3.229$ & $3.223$ \\
released, random & $7.556$ & $5.210$ & $4.192$ & $3.708$ & $3.482$ & $3.409$ & $3.376$ & $3.339$ & $3.324$ & $3.292$ & $3.286$ \\
dilated finetune, dilated & $7.557$ & $4.633$ & $3.685$ & $3.371$ & $3.256$ & $3.226$ & $3.212$ & $3.200$ & $3.194$ & $3.190$ & $3.188$ \\
dilated finetune, random & $7.557$ & $5.176$ & $4.146$ & $3.666$ & $3.439$ & $3.368$ & $3.333$ & $3.296$ & $3.280$ & $3.263$ & $3.257$ \\
uniform control, dilated & $7.556$ & $4.618$ & $3.664$ & $3.347$ & $3.231$ & $3.201$ & $3.186$ & $3.174$ & $3.167$ & $3.164$ & $3.161$ \\
uniform control, random & $7.556$ & $5.162$ & $4.130$ & $3.644$ & $3.415$ & $3.344$ & $3.309$ & $3.272$ & $3.256$ & $3.236$ & $3.230$ \\
\bottomrule
\end{tabular}
\end{table}
 
\begin{figure}[h]
\centering
\providecolor{C0}{HTML}{0072B2}\providecolor{C1}{HTML}{D55E00}\providecolor{C2}{HTML}{009E73}

\begin{tikzpicture}
\begin{axis}[width=0.40\columnwidth,height=0.26\columnwidth,scale only axis,xmode=log,log basis x=2,xtick={1,2,4,8,16,32,64,128},xticklabels={1,2,4,8,16,32,64,128},x tick label style={font=\fontsize{8}{9}\selectfont},y tick label style={font=\fontsize{8}{9}\selectfont},label style={font=\fontsize{8}{9}\selectfont},legend style={font=\fontsize{8}{9}\selectfont,draw=none,fill=none,at={(0.97,0.97)},anchor=north east},legend cell align=left,xlabel={decoding steps $T$ (spacing $b$)},ylabel={teacher-forced CE (nats/token)},tick align=outside,xtick pos=bottom,ytick pos=left]
\addplot[name path=releaseddL,draw=none,forget plot] coordinates {(1,7.51912) (2,4.62294) (4,3.67225) (8,3.34963) (16,3.22719) (24,3.21017) (32,3.19737) (48,3.20337) (64,3.18632) (96,3.19620) (128,3.19174)};
\addplot[name path=releaseddH,draw=none,forget plot] coordinates {(1,7.59417) (2,4.68663) (4,3.73381) (8,3.41225) (16,3.29050) (24,3.27211) (32,3.26240) (48,3.26609) (64,3.25004) (96,3.26022) (128,3.25508)};
\addplot[C0,opacity=0.15,forget plot] fill between[of=releaseddL and releaseddH];
\addplot[C0,mark=*,mark size=1pt,thick] coordinates {(1,7.55596) (2,4.65383) (4,3.70305) (8,3.38114) (16,3.25972) (24,3.24139) (32,3.23047) (48,3.23528) (64,3.21888) (96,3.22853) (128,3.22344)};

\addplot[name path=dilated_ftdL,draw=none,forget plot] coordinates {(1,7.52216) (2,4.60260) (4,3.65456) (8,3.34058) (16,3.22334) (24,3.19465) (32,3.18050) (48,3.16749) (64,3.16212) (96,3.15824) (128,3.15577)};
\addplot[name path=dilated_ftdH,draw=none,forget plot] coordinates {(1,7.59466) (2,4.66337) (4,3.71395) (8,3.40076) (16,3.28594) (24,3.25634) (32,3.24253) (48,3.23057) (64,3.22470) (96,3.22099) (128,3.21895)};
\addplot[C0!55!black,opacity=0.15,forget plot] fill between[of=dilated_ftdL and dilated_ftdH];
\addplot[C0!55!black,densely dotted,mark=*,mark size=1pt,thick] coordinates {(1,7.55749) (2,4.63289) (4,3.68465) (8,3.37077) (16,3.25597) (24,3.22633) (32,3.21211) (48,3.19981) (64,3.19444) (96,3.19028) (128,3.18845)};

\addplot[name path=uniform_controldL,draw=none,forget plot] coordinates {(1,7.51767) (2,4.58736) (4,3.63420) (8,3.31716) (16,3.19865) (24,3.16895) (32,3.15419) (48,3.14107) (64,3.13485) (96,3.13149) (128,3.12843)};
\addplot[name path=uniform_controldH,draw=none,forget plot] coordinates {(1,7.59419) (2,4.64913) (4,3.69382) (8,3.37742) (16,3.26067) (24,3.23105) (32,3.21631) (48,3.20377) (64,3.19761) (96,3.19343) (128,3.19152)};
\addplot[C2,opacity=0.15,forget plot] fill between[of=uniform_controldL and uniform_controldH];
\addplot[C2,mark=*,mark size=1pt,thick] coordinates {(1,7.55555) (2,4.61769) (4,3.66448) (8,3.34731) (16,3.23091) (24,3.20057) (32,3.18623) (48,3.17359) (64,3.16744) (96,3.16356) (128,3.16102)};

\addplot[C1,dashed,mark=o,mark size=1pt] coordinates {(1,7.55596) (2,5.20971) (4,4.19220) (8,3.70760) (16,3.48190) (24,3.40930) (32,3.37569) (48,3.33943) (64,3.32367) (96,3.29198) (128,3.28638)};

\legend{released (dilated),dilated-ft (dilated),uniform-control (dilated),released (random)}
\end{axis}
\end{tikzpicture}\hfill
\begin{tikzpicture}
\begin{axis}[width=0.40\columnwidth,height=0.26\columnwidth,scale only axis,xmode=log,log basis x=2,xtick={8,16,32,64,128},xticklabels={8,16,32,64,128},xmin=6,xmax=150,ymin=3.1,ymax=3.75,x tick label style={font=\fontsize{8}{9}\selectfont},y tick label style={font=\fontsize{8}{9}\selectfont},label style={font=\fontsize{8}{9}\selectfont},legend style={font=\fontsize{8}{9}\selectfont,draw=none,fill=none,at={(0.97,0.97)},anchor=north east},legend cell align=left,xlabel={decoding steps $T$ (spacing $b$)},ylabel={},tick align=outside,xtick pos=bottom,ytick pos=left]
\addplot[name path=releaseddL,draw=none,forget plot] coordinates {(1,7.51912) (2,4.62294) (4,3.67225) (8,3.34963) (16,3.22719) (24,3.21017) (32,3.19737) (48,3.20337) (64,3.18632) (96,3.19620) (128,3.19174)};
\addplot[name path=releaseddH,draw=none,forget plot] coordinates {(1,7.59417) (2,4.68663) (4,3.73381) (8,3.41225) (16,3.29050) (24,3.27211) (32,3.26240) (48,3.26609) (64,3.25004) (96,3.26022) (128,3.25508)};
\addplot[C0,opacity=0.15,forget plot] fill between[of=releaseddL and releaseddH];
\addplot[C0,mark=*,mark size=1pt,thick] coordinates {(1,7.55596) (2,4.65383) (4,3.70305) (8,3.38114) (16,3.25972) (24,3.24139) (32,3.23047) (48,3.23528) (64,3.21888) (96,3.22853) (128,3.22344)};

\addplot[name path=dilated_ftdL,draw=none,forget plot] coordinates {(1,7.52216) (2,4.60260) (4,3.65456) (8,3.34058) (16,3.22334) (24,3.19465) (32,3.18050) (48,3.16749) (64,3.16212) (96,3.15824) (128,3.15577)};
\addplot[name path=dilated_ftdH,draw=none,forget plot] coordinates {(1,7.59466) (2,4.66337) (4,3.71395) (8,3.40076) (16,3.28594) (24,3.25634) (32,3.24253) (48,3.23057) (64,3.22470) (96,3.22099) (128,3.21895)};
\addplot[C0!55!black,opacity=0.15,forget plot] fill between[of=dilated_ftdL and dilated_ftdH];
\addplot[C0!55!black,densely dotted,mark=*,mark size=1pt,thick] coordinates {(1,7.55749) (2,4.63289) (4,3.68465) (8,3.37077) (16,3.25597) (24,3.22633) (32,3.21211) (48,3.19981) (64,3.19444) (96,3.19028) (128,3.18845)};

\addplot[name path=uniform_controldL,draw=none,forget plot] coordinates {(1,7.51767) (2,4.58736) (4,3.63420) (8,3.31716) (16,3.19865) (24,3.16895) (32,3.15419) (48,3.14107) (64,3.13485) (96,3.13149) (128,3.12843)};
\addplot[name path=uniform_controldH,draw=none,forget plot] coordinates {(1,7.59419) (2,4.64913) (4,3.69382) (8,3.37742) (16,3.26067) (24,3.23105) (32,3.21631) (48,3.20377) (64,3.19761) (96,3.19343) (128,3.19152)};
\addplot[C2,opacity=0.15,forget plot] fill between[of=uniform_controldL and uniform_controldH];
\addplot[C2,mark=*,mark size=1pt,thick] coordinates {(1,7.55555) (2,4.61769) (4,3.66448) (8,3.34731) (16,3.23091) (24,3.20057) (32,3.18623) (48,3.17359) (64,3.16744) (96,3.16356) (128,3.16102)};

\addplot[C1,dashed,mark=o,mark size=1pt] coordinates {(1,7.55596) (2,5.20971) (4,4.19220) (8,3.70760) (16,3.48190) (24,3.40930) (32,3.37569) (48,3.33943) (64,3.32367) (96,3.29198) (128,3.28638)};

\end{axis}
\end{tikzpicture}
\caption{Forward sampling error against decoding steps: dilated order on the released model and both finetunes, and random order on the released model; values from \cref{tab:forward}. Left: all step counts; right: zoom on $T\ge8$.}
\label{fig:forwardce}
\end{figure}
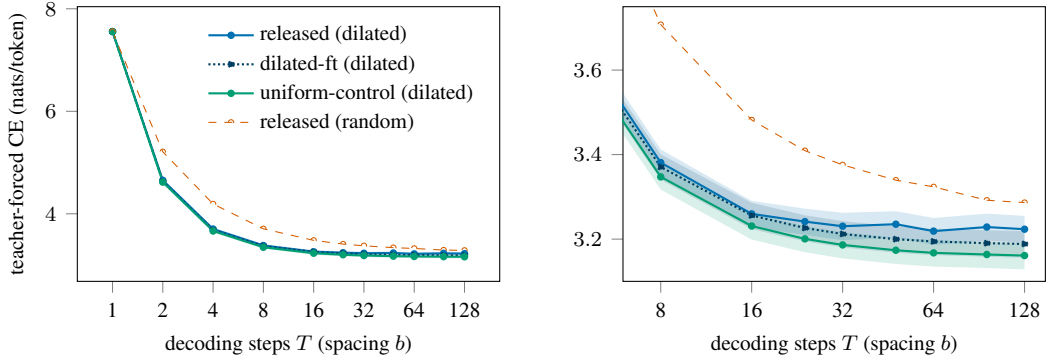

\paragraph{Results.} Dilated decoding lies below random order at every matched step count $T\ge2$ (\cref{tab:forward,fig:forwardce}), and below random order at twice the step count for $T\ge4$ on the released model and $T\ge8$ on the finetunes. The two schedules reach different floors of $3.223$ vs. $3.286$ at $T=128$. Dilated decoding saturates at $T_\delta=24$ (interval $[24,32]$) on the released model and at $32$ on both finetunes, against $96$ and $64$ for random order: inside the predicted band $[8,32]$, with two to four times fewer steps.
 
\subsubsection{Generation metrics}\label{app:generation}
 
We generate text from MDLM at $18$ step counts under three decoding orders and from LLaDA-8B at $8$ step counts under the dilated and random orders at matched per-step commit sizes, score the generations, and fit the step count at which quality saturates.
 
\paragraph{Setting.} We use the released MDLM checkpoint and LLaDA-8B-Base. We score generation with GPT-2-large; MAUVE also uses GPT-2-large features. Generation is unconditional at $N=1024$.  The MAUVE reference set is the first 512 windows of the eval stream. Because LLaDA-8B uses a different tokenizer, its outputs are detokenized, retokenized to GPT-2, and truncated to 1024 tokens for scoring.
 
\paragraph{Procedure.} At each step, we sample every committed position from the model's categorical with no truncation, in fp32 precision. The per-step commit counts are the dilated schedule's group sizes of $\lfloor N/T\rfloor$ at each order. We compare three decoding orders: dilated, random, and deterministic confidence-greedy \citep{chang2022maskgit}. For MDLM we evaluate the $14$ step counts $T\le128$ of its sweep plus $\{192,256,384,512\}$. The quality is the mean per-sequence NLL under GPT-2-large with out-of-vocab tokens clamped to EOS; we measure generation entropy and MAUVE \citep{pillutla2021mauve} (mauve-text $0.40$). We use the saturation estimator
 
\begin{equation}\label{eq:genfit}
\hat T^*=2^{(a-\ell_{\mathrm{floor}})/c},\qquad (a,c,\ell_{\mathrm{floor}})=\arg\min\sum_{T\leq 128}\left(\ell(T)-\max(a-c\log_2T, \ell_{\mathrm{floor}})\right)^2
\end{equation}
where the interval comes from resampling the points and refitting. LLaDA-8B generates at eight counts in $\{1,\cdots, 128\}$.
 
\begin{table}[h]
\caption{Generation quality on the released MDLM checkpoint at every
step count of the sweep and the extended grid: mean per-sequence NLL
under GPT-2-large (nats per token, $95\%$ bootstrap intervals over
sequences) and MAUVE against the held-out reference, for the three
decoding orders at matched per-step commit sizes.}
\label{tab:gen}
\centering\footnotesize\setlength{\tabcolsep}{3pt}
\begin{tabular}{@{}lcccccc@{}}
\toprule
& \multicolumn{2}{c}{dilated} & \multicolumn{2}{c}{random} & \multicolumn{2}{c}{confidence} \\
\cmidrule(lr){2-3}\cmidrule(lr){4-5}\cmidrule(lr){6-7}
$T$ & NLL & MAUVE & NLL & MAUVE & NLL & MAUVE \\
\midrule
$1$ & $8.460$ $[8.45,8.47]$ & $0.005$ & $8.461$ $[8.45,8.47]$ & $0.006$ & $8.460$ $[8.45,8.47]$ & $0.006$ \\
$2$ & $7.420$ $[7.41,7.43]$ & $0.022$ & $8.132$ $[8.12,8.14]$ & $0.007$ & $8.557$ $[8.55,8.57]$ & $0.005$ \\
$3$ & $6.636$ $[6.62,6.65]$ & $0.068$ & $7.841$ $[7.83,7.85]$ & $0.007$ & $8.554$ $[8.54,8.56]$ & $0.005$ \\
$4$ & $6.116$ $[6.10,6.13]$ & $0.158$ & $7.583$ $[7.57,7.60]$ & $0.011$ & $8.539$ $[8.53,8.55]$ & $0.006$ \\
$6$ & $5.652$ $[5.63,5.67]$ & $0.341$ & $7.147$ $[7.13,7.17]$ & $0.007$ & $8.512$ $[8.50,8.53]$ & $0.005$ \\
$8$ & $5.387$ $[5.37,5.41]$ & $0.443$ & $6.698$ $[6.67,6.72]$ & $0.020$ & $8.480$ $[8.46,8.50]$ & $0.006$ \\
$12$ & $5.287$ $[5.26,5.31]$ & $0.580$ & $6.123$ $[6.10,6.15]$ & $0.010$ & $8.400$ $[8.38,8.42]$ & $0.006$ \\
$16$ & $5.014$ $[4.99,5.04]$ & $0.512$ & $5.800$ $[5.77,5.83]$ & $0.015$ & $8.316$ $[8.29,8.34]$ & $0.006$ \\
$24$ & $4.920$ $[4.89,4.95]$ & $0.615$ & $5.468$ $[5.44,5.49]$ & $0.034$ & $8.105$ $[8.07,8.14]$ & $0.004$ \\
$32$ & $4.883$ $[4.85,4.92]$ & $0.439$ & $5.310$ $[5.29,5.34]$ & $0.016$ & $7.920$ $[7.88,7.96]$ & $0.005$ \\
$48$ & $4.922$ $[4.89,4.96]$ & $0.506$ & $5.081$ $[5.05,5.11]$ & $0.025$ & $7.782$ $[7.74,7.83]$ & $0.006$ \\
$64$ & $4.835$ $[4.79,4.88]$ & $0.251$ & $4.947$ $[4.92,4.97]$ & $0.016$ & $7.626$ $[7.57,7.68]$ & $0.006$ \\
$96$ & $4.894$ $[4.84,4.94]$ & $0.257$ & $4.906$ $[4.88,4.93]$ & $0.024$ & $7.390$ $[7.32,7.46]$ & $0.005$ \\
$128$ & $4.785$ $[4.73,4.83]$ & $0.167$ & $4.771$ $[4.74,4.80]$ & $0.013$ & $7.096$ $[7.00,7.19]$ & $0.008$ \\
$192$ & $4.801$ $[4.73,4.87]$ & $0.346$ & $4.747$ $[4.72,4.77]$ & $0.011$ & $6.675$ $[6.55,6.80]$ & $0.007$ \\
$256$ & $4.579$ $[4.51,4.66]$ & $0.083$ & $4.734$ $[4.71,4.76]$ & $0.016$ & $5.990$ $[5.83,6.15]$ & $0.009$ \\
$384$ & $4.435$ $[4.34,4.54]$ & $0.074$ & $4.690$ $[4.67,4.72]$ & $0.014$ & $5.306$ $[5.12,5.49]$ & $0.009$ \\
$512$ & $4.079$ $[3.96,4.20]$ & $0.021$ & $4.661$ $[4.63,4.69]$ & $0.011$ & $4.127$ $[3.92,4.33]$ & $0.013$ \\
\bottomrule
\end{tabular}
\end{table}
 
\paragraph{Results.} Every order's NLL reaches its minimum at $512$ steps while dilated's MAUVE peaks at $0.61$ at $24$ steps and falls through $0.25$ at $64$ and $0.17$ at $128$ to $0.02$ at $512$. The confidence order's MAUVE never exceeds $0.013$; generation entropy falls together with MAUVE, and at $512$ steps dilated and the confidence order have overlapping NLL intervals ($4.079$ against $4.127$). LLaDA-8B improves monotonically across its tested range without observed saturation. The dilated fit saturates at $\hat T^*=9.7$ steps, inside the predicted band $[8,32]$ and before random order's $44.7$, so both generation-side predictions hold.
 
\subsection{Loop-count saturation}\label{app:loop}
 
We train a looped variant of MDLM from scratch, evaluate held-out loss at every loop count at three training checkpoints, and fit the loop count at which loss saturates. This experiment is qualitative and incomplete: \cref{cor:looped} concerns the idealized message-passing class of \cref{sec:looped} and suggests this saturation point grows as training reduces the achievable excess.
 
\paragraph{Setting.} We train a looped variant of MDLM with $12$ stored blocks at hidden size $768$ with a $128$-dimensional conditioning pathway, of which block $0$ runs once as the head, blocks $1$ and $2$ are shared and applied $\Lambda$ times, and blocks $3$ to $11$ run once; the model has $170$M parameters, context $1024$, and the GPT-2 vocabulary with a mask token. We train on the OWT training region and evaluate on $1024$-token windows from the evaluation stream.
 
\paragraph{Procedure.} We trained for 4B tokens (7630 steps of 512 sequences) with checkpoints at 1, 2, and 4B, deliberately short of convergence to track behavior over training. The loop count was drawn uniformly on $\{1,2,\cdots, 12\}$ per microbatch and the loss is the ELBO's $1/t$-weighted masked CE with $t$ clamped at $10^{-3}$. We used the same masks across choices of $\Lambda$. We fit the saturation point
\begin{equation}\label{eq:loopfit}
\hat\Lambda^*=\frac{a-\ell_{\mathrm{floor}}}{c},\qquad (a,c,\ell_{\mathrm{floor}})=\arg\min\sum_{\Lambda}\left(\ell(\Lambda)-\max(a-c\Lambda, \ell_{\mathrm{floor}})\right)^2
\end{equation}
 
\paragraph{Results.} \Cref{tab:loop} gives the fitted saturation points of $2.419\to2.528\to2.621$. The extrapolated 4B $\hat{\Lambda}^*$ of 2.637 $[2.60,2.67]$ contains the measured $2.621$ $[2.608,2.635]$. Excluding $\Lambda=1$ moves all three points right by close to 1 with ordering and spacing preserved; past the loop count minimizer for each checkpoint, the loss rises by 0.028. The saturation point moves right with training.
 
\begin{table}[h]
\caption{Loop-count saturation at the three checkpoints: the fitted
saturation point on the full grid, the refit excluding
$\Lambda=1$, and the loop count at which held-out loss is lowest.}
\label{tab:loop}
\centering\small
\begin{tabular}{@{}lccc@{}}
\toprule
checkpoint & $\hat\Lambda^*$ (all counts) & $\hat\Lambda^*$ ($\Lambda\ge2$) & loss minimum at \\
\midrule
$1$B & $2.419$ & $3.16$ & $\Lambda=6$ \\
$2$B & $2.528$ & $3.42$ & $\Lambda=6$ \\
$4$B & $2.621$ $[2.608,2.635]$ & $3.61$ & $\Lambda=7$ \\
\bottomrule
\end{tabular}
\end{table}

\end{document}